\documentclass[10pt,journal,compsoc]{IEEEtran}
\ifCLASSINFOpdf
\else
\fi
\usepackage{amsmath,amsfonts}
\usepackage{algorithmic}
\usepackage{algorithm}
\usepackage{array}

\usepackage[caption=false,font=normalsize,labelfont=sf,textfont=sf]{subfig}
\usepackage{caption}
\usepackage{textcomp}
\usepackage{stfloats}
\usepackage{url}
\usepackage{epigraph} 
\usepackage{pifont}
\usepackage{array} 
\usepackage{mydef}

\usepackage[normalem]{ulem}

\usepackage{tikz}
\usepackage{forest}
\usetikzlibrary{trees,positioning,shapes,shadows,arrows.meta}

\usepackage{verbatim}
\usepackage{graphicx}
\usepackage{cite}
\usepackage[T1]{fontenc}
\usepackage[utf8]{inputenc}
\usepackage{makecell}
\usepackage{multirow}
\usepackage{xcolor}
\usepackage{enumitem}
\usepackage{ragged2e}

\usepackage[]{mdframed}
\usepackage{xurl}
\usepackage[bookmarks=true, hidelinks, breaklinks,colorlinks,citecolor=blue]{hyperref}

\usepackage[skins]{tcolorbox}

\newcommand{\added}[1]{#1}
\newenvironment{addedblock}{}{}
\newcommand{\deleted}[1]{}

\def\ie{{\em i.e.}}
\def\eg{{\em e.g.}}

\begin{document}

\title{Replication in 
Visual Diffusion Models: \\
A Survey and Outlook}

\author{Wenhao Wang, Yifan Sun, Zongxin Yang, Zhengdong Hu, Zhentao Tan, Yi Yang$^*$, Fellow, IEEE
\IEEEcompsocitemizethanks{
\IEEEcompsocthanksitem Wenhao Wang and Zhengdong Hu are with the Australian Artificial Intelligence Institute, University of Technology Sydney, Sydney, New South Wales, Australia. e-mail: wangwenhao0716@gmail.com and huzhengdongcs@gmail.com.
\IEEEcompsocthanksitem Yifan Sun is with Baidu Inc., Beijing, China. e-mail: sunyf15@tsinghua.org.cn.
\IEEEcompsocthanksitem Zhentao Tan, Zongxin Yang and Yi Yang are with the College of Computer Science and Technology, Zhejiang University, Hangzhou, Zhejiang, China. e-mail: tanzhentao990514@gmail.com, zongxinyang1996@gmail.com and yangyics@zju.edu.cn.
\IEEEcompsocthanksitem Yi Yang is the corresponding author.
}

}



\IEEEtitleabstractindextext{%
\begin{abstract}
\justifying
Visual diffusion models have revolutionized the field of creative AI, producing high-quality and diverse content. 
However, they inevitably memorize training images or videos, subsequently \textit{replicating} their concepts, content, or styles during inference. This phenomenon raises significant concerns about privacy, security, and copyright within generated outputs.
In this survey, we provide the first comprehensive review of replication in visual diffusion models, marking a novel contribution to the field by systematically categorizing the existing studies into unveiling, understanding, and mitigating this phenomenon.
Specifically, \textit{unveiling} mainly refers to the methods used to detect replication instances. \textit{Understanding} involves analyzing the underlying mechanisms and factors that contribute to this phenomenon. \textit{Mitigation} focuses on developing strategies to reduce or eliminate replication. Beyond these aspects, we also review papers focusing on its real-world influence. For instance, in the context of healthcare, replication is critically worrying due to privacy concerns related to patient data.
Finally, the paper concludes with a discussion of the ongoing challenges, such as the difficulty in detecting and benchmarking replication, and outlines future directions including the development of more robust mitigation techniques.
By synthesizing insights from diverse studies, this paper aims to equip researchers and practitioners with a deeper understanding at the intersection between AI technology and social good.
We release this project at https://github.com/WangWenhao0716/Awesome-Diffusion-Replication.

\end{abstract}

\begin{IEEEkeywords}
Replication, Visual Diffusion Models, AI for Social Good, AI Security
\end{IEEEkeywords}}

\maketitle
\IEEEdisplaynontitleabstractindextext

%
\IEEEpeerreviewmaketitle

\section{Introduction}

\IEEEPARstart{V}{isual} diffusion models represent a significant advancement in the field of generative modeling, particularly for image synthesis tasks. These models leverage the concept of diffusion, a process inspired by statistical physics, to generate images from random noise \cite{sohl2015deep,ho2020denoising}. Compared to traditional Generative Adversarial Networks (GAN) \cite{goodfellow2014generative} and Variational Autoencoders (VAE) \cite{kingma2013auto}, visual diffusion models excel in producing high-quality, diverse, and stable images. Famous visual diffusion models include OpenAI’s DALL-E \cite{ramesh2021zero,ramesh2022hierarchical,betker2023improving}, Stability AI’s Stable Diffusion \cite{rombach2022high,podell2023sdxl,esser2024scaling}, Google’s Imagen \cite{saharia2022photorealistic}, and Baidu's ERNIE-ViLG \cite{zhang2021ernie,feng2023ernie}, drawing widespread attention from researchers, practitioners, and enthusiasts.\par 
Visual diffusion models have a broad range of real-world applications across various industries. In the entertainment sector, these models are utilized for creating highly realistic visual effects \cite{po2023state}, animations \cite{guo2023animatediff}, and virtual environments in movies and video games \cite{shen2024context}, significantly reducing production costs and time. In the field of design and fashion, they aid in generating new styles, patterns, and prototypes, fostering innovation and creativity \cite{cao2023difffashion,baldrati2023multimodal,sun2023sgdiff}. Marketing and advertising benefit from these models through the creation of visually appealing and customized content that enhances consumer engagement \cite{digital4010013}. Additionally, in healthcare, visual diffusion models assist in medical imaging by enhancing the quality of diagnostic images \cite{asgariandehkordi2023deep,wang2024implicit} and creating synthetic data for research and training purposes \cite{ali2022spot,pinaya2022brain}. 
The image-generating AI market is estimated to be valued at around $349.6$ million in 2023 and is expected to grow to approximately $1,081.2$ million by 2030 \cite{grandview2024}.\par

\begin{figure}[t]
    \centering
    \includegraphics[width=0.47\textwidth]{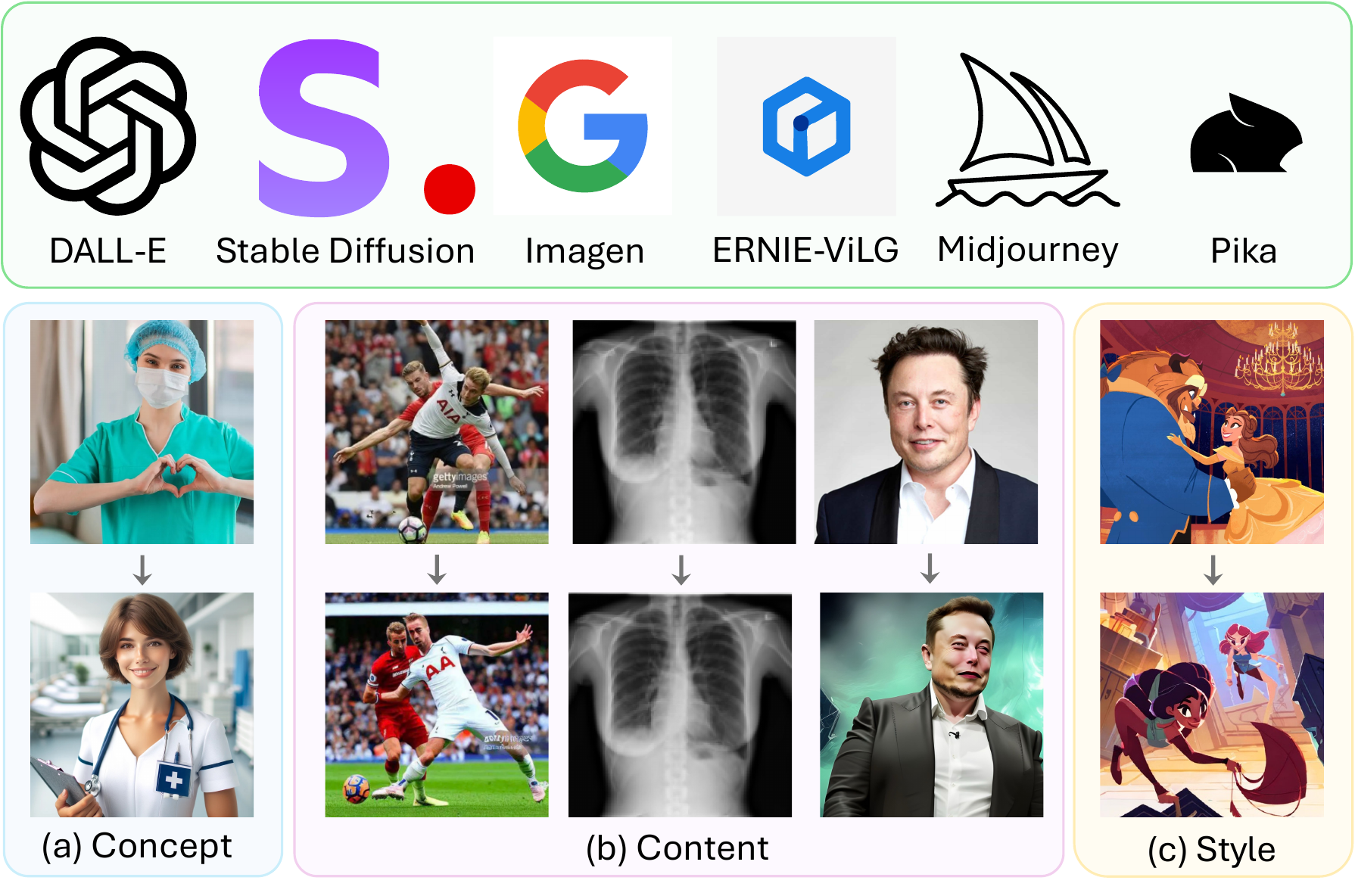}
    \vspace{-2mm}
    \caption{During training, visual diffusion models memorize the training images and \textit{replicate} their concepts, content, or styles during the inference stage. For instance, they can replicate (a) a biased concept of ``nurses are female'', (b) copyrighted content from Getty Images, private content from patient X-ray films, and facial portrait from Elon Musk, and (c) unique stylistic elements from a contemporary artist, Hollie Mengert.}
    \label{fig:teasor}
    \vspace{-4mm}
\end{figure}

To achieve such outstanding performance and broad applications, visual diffusion models highly rely on extensive web data, such as LAION-5B \cite{schuhmann2022laion}, for training. However, this data encompasses several significant issues: First, at the \textit{concept} level, the training data often contains biased gender \cite{anand2023identifying} and culture \cite{struppek22homoglyphs}, racial representations \cite{luccioni2023stable}, and Not Safe For Work (NSFW) materials \cite{schramowski2023safe}. Second, at the \textit{content} level, web data includes a substantial amount of copyrighted images \cite{somepalli2023understanding}, medical images containing patient information \cite{kazerouni2023diffusion}, and photos of politicians or celebrities \cite{chen2023text}. Third, at the \textit{style} level, data may include works characterized by unique stylistic elements from contemporary artists \cite{somepalli2024measuring,wang2024AnyPattern}. These issues lead to some generated images exhibiting unfair outcomes, inappropriate content, ethical risks, and copyright infringement \cite{wang2023security}, thereby negatively impacting the widespread applications of visual diffusion models.\par

Fundamentally, as shown in Fig. \ref{fig:teasor}, this problem comes from an inevitable and important phenomenon in current visual diffusion models, \textit{i.e.}, during training, these models memorize the training images and \textit{replicate} their concepts, content, or styles during the inference stage. Currently, an increasing amount of research is being conducted to discuss this \textit{replication} phenomenon. However, there is a lack of surveys that specifically focus on replication in visual diffusion models. In this survey, we provide the first comprehensive review of replication in visual diffusion models, which not only systematically investigates this research topic but also potentially benefits the improvement of model safety and ethical standards in the real world. \par 
Our survey systematically introduces the concept of replication in visual diffusion models from three perspectives: \textit{unveiling}, \textit{understanding}, and \textit{mitigation}. \textit{Unveiling} involves identifying and exposing replication through techniques such as similarity retrieval \cite{somepalli2023diffusion,somepalli2024measuring}, membership inference \cite{duan2023diffusion,kong2023efficient}, and prompting \cite{leotta2023not,wen2024hard}. \textit{Understanding} explores the mechanisms behind replication, including factors like data duplication \cite{somepalli2023understanding,naseh2023memory} and inappropriate training methods \cite{yi2023generalization,zhang2023emergence}. \textit{Mitigation} discusses strategies to minimize replication, such as differential privacy \cite{dockhorn2023differentially,lyu2023differentially}\added{,\cite{tsai2025dplora}}, data deduplication \cite{webster2023duplication,abbas2023semdedup}, machine unlearning \cite{gandikota2023erasing,kumari2023ablating}\added{,\cite{hasanaliyev2025dataunlearning,george2025illusion}}, and \added{inference-time anti-replication guidance \cite{li2025loyaldiffusion}}. Lastly, we explore the influence of replication in the real world, including regulation \cite{zirpoli2023generative,lee2023ai}, art \cite{aiart,crawford2022legal}, society \cite{luccioni2023stable,perera2023analyzing}, and healthcare \cite{dar2023investigating,fernandez2023privacy}.
An overview of this survey is available at Fig. \ref{Fig: overview}.

This survey makes the following contributions:
\begin{enumerate}[leftmargin=*]
 \item This is the first survey that systematically reviews the concept of replication in visual diffusion models. We innovatively discuss this phenomenon from the perspectives of unveiling, understanding, mitigation, and its influence in the real-world.
  \item We provide a brief overview of visual diffusion models, including their categorization, theoretical foundations, and functionalities. We then formally introduce the term replication within this context, providing a concise definition and understanding of its meaning.
  \item By pointing out the inadequacies of current methods and the challenges existing in replication, we provide a roadmap for future research, such as developing more accurate and efficient unveiling methods and creating more robust mitigation strategies.
\end{enumerate}


\tikzstyle{leaf1}=[draw=none, 
    rounded corners,minimum height=1.2em,
    edge=black!10, 
    text opacity=1, align=center,
    fill opacity=.3,  text=black,font=\small,
    inner xsep=2pt, inner ysep=3.6pt,
    ]
\tikzstyle{leaf2}=[draw=none, 
    rounded corners,minimum height=1.2em,
    edge=black!10, 
    text opacity=1, align=center,
    fill opacity=.5,  text=black,font=\small,
    inner xsep=2pt, inner ysep=3.6pt,
    ]
\tikzstyle{leaf3}=[draw=none, 
    rounded corners,minimum height=1.2em,
    edge=black!10, 
    text opacity=1, align=center,
    fill opacity=.8,  text=black,font=\small,
    inner xsep=2pt, inner ysep=3.8pt,
    ]
\tikzstyle{leaf4}=[draw=none, 
    rounded corners,minimum height=1.2em,
    edge=black!10, 
    text opacity=1, align=center,
    fill opacity=1,  text=black,font=\small,
    inner xsep=2pt, inner ysep=3.8pt,
    ]
    
\tikzstyle{leaf5}=[draw=none, 
    rounded corners,minimum height=1.2em,
    edge=black!10, 
    text opacity=1, align=center,
    fill opacity=1,  text=black,font=\small,
    inner xsep=2pt, inner ysep=3.8pt,
    ]

The remainder of this survey is organized as follows: In Section \ref{related}, we highlight the differences between our survey and existing ones. In Section \ref{background}, we briefly introduce visual diffusion models and define the phenomenon of replication. Subsequently, we summarize unveiling, understanding, and mitigation in Sections \ref{Unveiling}, \ref{Understanding}, and \ref{Mitigation}, respectively. Additionally, in Section \ref{Medical}, we review papers that focus on the influence of replication in the real world. Finally, we present the current challenges and future directions in Section \ref{Challenges} and conclude the survey in Section \ref{Conclusion}.



\section{Related Works}\label{related}
\noindent\textbf{Diffusion models in vision.} Existing surveys on diffusion models, such as \cite{croitoru2023diffusion,zhang2023text,yang2023diffusion,cao2024survey}, provide comprehensive overviews of various diffusion modeling techniques and their applications in computer vision. These surveys categorize diffusion models, discuss their theoretical foundations, and highlight their performance in tasks like image synthesis and data augmentation. In contrast, our survey uniquely focuses on the critical issue of replication within diffusion models. We systematically explore this phenomenon through the lenses of unveiling, understanding, and mitigation, thereby filling a gap between general diffusion model overviews and the specific challenge of replication.

\noindent\textbf{Safety of diffusion models.} Existing surveys on the safety of diffusion models often address issues such as bias, misinformation, privacy concerns, and copyright protection. For instance, \cite{lin2024detecting} emphasizes the critical need to identify AI-generated content to prevent its misuse and potential societal disruptions. \cite{wang2023security} explores privacy risks associated with generative AI and highlights the importance of robust detection and authentication. Additionally, \cite{fan2023trustworthiness} and \cite{chen2023challenges} investigate the broader ethical implications and technical challenges of ensuring the integrity and trustworthiness of AI-generated content, including the use of privacy-preserving techniques and blockchain for content verification. Furthermore, \cite{ren2024copyright} addresses the legal and technical challenges of protecting intellectual property rights in the context of AI-generated works, emphasizing the need to identify and verify copyrighted content.

\textbf{In contrast}, while our survey also falls under the safety of diffusion models, we specifically target the replication phenomenon within visual diffusion models. This focus is unique compared to existing surveys: while these surveys emphasize detection and mitigation of AI-generated content to prevent misuse and ensure ethical deployment, our survey goes deeper into the intrinsic properties of diffusion models related to replication. This distinction not only complements existing studies but also provides a more granular understanding of the safety concerns associated with visual diffusion models.

\noindent\textbf{Replication in large language models.} The replication phenomenon in large language models (LLMs) have been extensively studied in recent literature. Works such as \cite{hartmann2023sok} explore the implications of memorization for privacy, security, and copyright. Similarly, the survey \cite{ishihara2023training} provides a comprehensive overview of methods for extracting training data from LLMs and discusses the inherent challenges in mitigating these risks. Our survey differentiates itself by focusing specifically on visual diffusion models, filling this gap in the current literature.

\begin{addedblock}
\noindent\textbf{Comparison with related surveys.} To further clarify the position and contributions of our survey, Table~\ref{tab:survey_comparison} provides a structured comparison with representative related works. Our survey is the \textit{only} one that (i) concentrates exclusively on the replication phenomenon in visual diffusion models, (ii) covers all three semantic levels (concept, content, style), (iii) provides a formal threshold-based definition, and (iv) includes a dedicated section on real-world influence spanning regulation, art, society, and healthcare.

\begin{table*}[t]
\caption{Comparison of our survey with related works. \checkmark~indicates that the topic is covered; $\circ$ indicates partial coverage; $\times$ indicates not covered.}
\label{tab:survey_comparison}
\setlength{\tabcolsep}{3.8pt}
\renewcommand{\arraystretch}{1}
\begin{tabular*}{\textwidth}{@{\extracolsep{\fill}}lccccccc}
\hline
\textbf{Survey} & \textbf{Visual DMs} & \textbf{Replication} & \textbf{3 Levels} & \textbf{Formal Def.} & \textbf{Unveil/Understand/Mitigate} & \textbf{Real-World Influence} & \textbf{Video/3D} \\
\hline
Croitoru~\textit{et al.}~\cite{croitoru2023diffusion} & \checkmark & $\times$ & $\times$ & $\times$ & $\times$ & $\times$ & $\circ$ \\
Cao~\textit{et al.}~\cite{cao2024survey} & \checkmark & $\times$ & $\times$ & $\times$ & $\times$ & $\times$ & $\circ$ \\
Wang~\textit{et al.}~\cite{wang2023security} & \checkmark & $\circ$ & $\times$ & $\times$ & $\circ$ & $\circ$ & $\times$ \\
Fan~\textit{et al.}~\cite{fan2023trustworthiness} & \checkmark & $\circ$ & $\times$ & $\times$ & $\circ$ & $\circ$ & $\times$ \\
Hartmann~\textit{et al.}~\cite{hartmann2023sok} & $\times$ & \checkmark & $\times$ & $\circ$ & $\circ$ & $\times$ & $\times$ \\
Ishihara~\textit{et al.}~\cite{ishihara2023training} & $\times$ & \checkmark & $\times$ & $\times$ & $\circ$ & $\times$ & $\times$ \\
\textbf{Ours} & \checkmark & \checkmark & \checkmark & \checkmark & \checkmark & \checkmark & \checkmark \\
\hline
\end{tabular*}
\vspace{-3mm}
\end{table*}
\end{addedblock}

\section{Background}\label{background}
In this section, we provide an overview of visual diffusion models and formally define the replication phenomenon.

\subsection{Visual Diffusion Models}

\noindent\textbf{Categorization and theoretical foundations.} Diffusion models are typically categorized into three main types: denoising diffusion probabilistic models (DDPMs) \cite{ho2020denoising}, noise-conditioned score networks (NCSNs)  \cite{song2019generative}, and stochastic differential equations (SDEs) \cite{song2020score}.

\textit{Denoising Diffusion Probabilistic Models (DDPMs)}:
   DDPMs add Gaussian noise to the data in a forward process and learn to reverse this process to denoise the data. The forward process is defined as:
  \begin{equation}
   q(\mathbf{x}_t | \mathbf{x}_{t-1}) = \mathcal{N}(\mathbf{x}_t; \sqrt{\alpha_t} \mathbf{x}_{t-1}, (1 - \alpha_t) \mathbf{I}),
\end{equation}
   where \( \alpha_t \) is a noise schedule parameter. The reverse process is:
  \begin{equation}
   p_\theta(\mathbf{x}_{t-1} | \mathbf{x}_t) = \mathcal{N}(\mathbf{x}_{t-1}; \mathbf{\mu}_\theta(\mathbf{x}_t, t), \sigma_t^2 \mathbf{I}),
\end{equation}
   with \(\mathbf{\mu}_\theta\) being predicted by a neural network.\par

\textit{Noise-Conditioned Score Networks (NCSNs)}: NCSNs estimate the score function, the gradient of the log density of the data, to denoise the data. The forward process introduces noise, and the model learns to predict the score:
\begin{equation}
   \mathbf{s}_\theta(\mathbf{x}_t, t) \approx \nabla_{\mathbf{x}_t} \log p(\mathbf{x}_t).
\end{equation}
   The reverse process uses Langevin dynamics to generate new samples:
\begin{equation}
   \mathbf{x}_{t+1} = \mathbf{x}_t + \frac{\epsilon^2}{2} \mathbf{s}_\theta(\mathbf{x}_t, t) + \epsilon \mathbf{z}, \quad \mathbf{z} \sim \mathcal{N}(0, \mathbf{I}),
\end{equation}
   where \(\epsilon\) is a step size parameter.

\textit{Stochastic Differential Equations (SDEs)}:
   SDEs generalize the diffusion process using continuous-time dynamics. The forward process can be described by an SDE:
\begin{equation}
   d\mathbf{x}_t = \mathbf{f}(\mathbf{x}_t, t) dt + g(t) d\mathbf{w}_t,
\end{equation}
   where \(\mathbf{w}_t\) is a standard Wiener process. The reverse-time SDE is used to generate samples:
\begin{equation}
   d\mathbf{x}_t = [\mathbf{f}(\mathbf{x}_t, t) - g(t)^2 \nabla_{\mathbf{x}_t} \log p_t(\mathbf{x}_t)] dt + g(t) d\mathbf{\hat{w}}_t,
\end{equation}
   where \(d\mathbf{\hat{w}}_t\) is the reverse-time Wiener process.\par

\noindent\textbf{Functionalities.} Visual diffusion models exhibit a broad range of functionalities, including 
storytelling \cite{rahman2023make,liu2024intelligent,song2024causal}, 
virtual try-on \cite{morelli2023ladi,gou2023taming,kim2024stableviton}, 
drag edit \cite{mou2023dragondiffusion,ling2024freedrag,shi2024dragdiffusion}, 
diffusion inversion \cite{mokady2023null,ju2023direct,zhang2023inversion}, 
text-guided editing \cite{hertz2022prompt,parmar2023zero,brooks2023instructpix2pix}, 
T2Iaugmentation \cite{chefer2023attend,hong2023improving,ge2023expressive}, 
spatial control \cite{zeng2023scenecomposer, li2023gligen,feng2022training}, 
image translation \cite{meng2021sdedit, xu2024cyclenet,qi2024deadiff}, 
inpainting \cite{yang2023paint, avrahami2022blended,avrahami2023blended}, 
layout generation \cite{inoue2023layoutdm, weng2024desigen}, 
super resolution \cite{yue2024resshift, saharia2022image}, 
video generation \cite{khachatryan2023text2video, nikankin2023sinfusion}, 
and video editing \cite{qi2023fatezero, bar2022text2live}, showing their versatility and applicability across diverse domains. 
However, at the same time, visual diffusion models also pose potential threats to this wide range of functionalities through the replication of their training data. This underscores the necessity of our survey, which provides a comprehensive review of this phenomenon, aiming to enhance model safety and ethical standards.

\subsection{Replication}
\begin{addedblock}
\noindent\textbf{Definition.} Let $\mathcal{T}=\left\{x_1, x_2, \ldots, x_n\right\}$ denote a training set of $n$ samples. A diffusion model trained on this set is denoted as $f_{\mathcal{T}}$. During the inference phase, the model generates a set of $m$ data points denoted as $\mathcal{G}=\left\{\hat{x}_1, \ldots, \hat{x}_m\right\}$. Given a feature extractor $\phi$ and a distance metric $d$ defined in the induced feature space, we say that a trained diffusion model $f_\mathcal{T}$ \textit{replicates} a training sample $x_i \in \mathcal{T}$ via a generated output $\hat{x}_j \in \mathcal{G}$ if and only if
\begin{equation}
    d\!\left(\phi(\hat{x}_j),\, \phi(x_i)\right) < \tau,
\end{equation}
where $\tau > 0$ is a task-specific threshold calibrated on a labeled replication dataset. We say that $f_\mathcal{T}$ \textit{replicates its training set} if such a pair $(x_i, \hat{x}_j)$ exists.

\noindent\textbf{Remark on the threshold $\tau$.} Because replication occurs at multiple semantic levels, $\phi$ and $\tau$ are defined separately for each level:
\begin{itemize}[leftmargin=*]
 \item \textit{Content level}: $\phi$ is a copy-detection backbone (\textit{e.g.}, SSCD \cite{pizzi2022self}); $d$ is cosine distance; $\tau_\text{content}$ is calibrated on labeled near-duplicate pairs to maximize the F1 score.
 \item \textit{Style level}: $\phi$ extracts Gram-matrix or style-space features \cite{somepalli2024measuring,wang2024AnyPattern}; $\tau_\text{style}$ is calibrated on artist-attributed style pairs to capture stylistic near-replication without requiring identical content.
 \item \textit{Concept level}: $\phi$ is a semantic encoder (\textit{e.g.}, CLIP \cite{radford2021learning}); $\tau_\text{concept}$ is calibrated on concept-level annotations to capture high-level semantic similarity, including biased or inappropriate conceptual replications.
\end{itemize}
The distance metric $d$ is a function defined on the feature space $\mathcal{F}$: $\mathcal{F} \times \mathcal{F} \rightarrow \mathbb{R}_{\geq 0}$, satisfying for all $u, v, w \in \mathcal{F}$:
\begin{itemize}[leftmargin=*]
 \item $d(u, v)=0 \iff u=v$ (identity of indiscernibles);
 \item $d(u, v)=d(v, u)$ (symmetry);
 \item $d(u, w)\leq d(u, v)+d(v, w)$ (triangle inequality).
\end{itemize}
\end{addedblock}

\section{Unveiling}\label{Unveiling}
\begin{figure*}[t]
    \centering
    \includegraphics[width=1\textwidth]{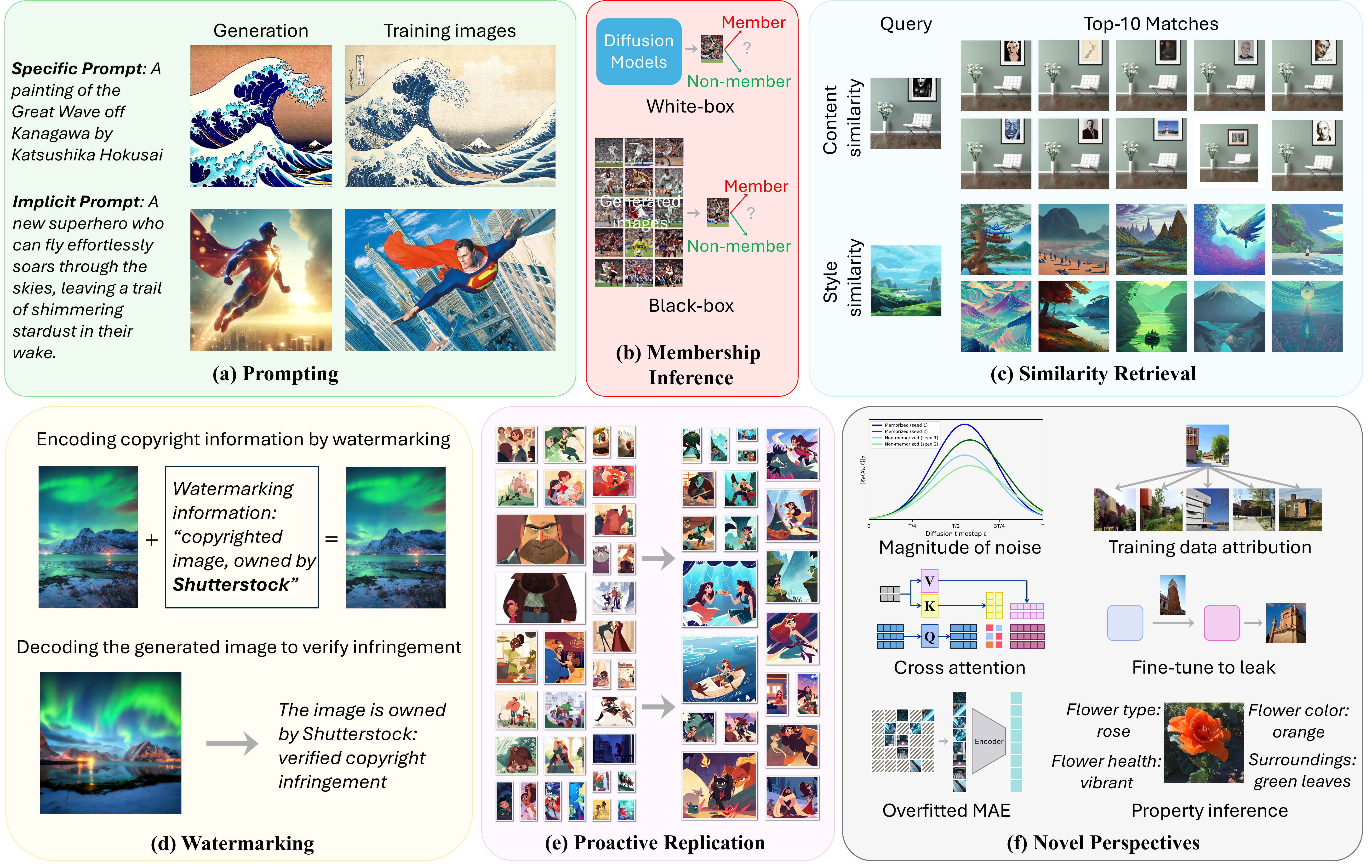}
    \vspace{-6mm}
    \caption{Illustrations of different methods for unveiling replication in visual diffusion models.}
    \label{Fig: unveil}
    \vspace{-4mm}
\end{figure*}

In this section, we focus on the first aspect of our survey on replication in visual diffusion models, which is \textit{unveiling}. Unveiling \cite{amer2023ai,RePEc,m2024challenges,rando2022red} refers to the process of uncovering the phenomenon of replication, either manually or through the use of specially-designed machine learning models.
As shown in Fig. \ref{Fig: overview}, we organize the unveiling of replication into 6 categories, \textit{i.e.}, prompting, membership inference, similarity retrieval, proactive replication, watermarking, and novel perspectives.  Fig. \ref{Fig: unveil} illustrates these categories.

\subsection{Prompting}
These articles investigate how prompting can reveal replication in visual diffusion models. As shown in Fig. \ref{Fig: unveil} (a): by using \textit{specific} prompts\added{, \ie, prompts deliberately designed to elicit memorized content by including verbatim or near-verbatim training captions}, researchers can generate images that closely resemble the model’s training data; beyond that, some papers show visual diffusion models may replicate learned copyrighted images \textit{implicitly}\added{, \ie, through prompts that indirectly trigger replication by describing concepts strongly associated with specific training images}. 

\noindent\textbf{Specific.} Specific prompts are carefully chosen phrases or descriptions from researchers to test whether visual diffusion models can replicate. For instance, \cite{somepalli2023diffusion,carlini2023extracting,webster2023reproducible,naseh2023understanding} employ specific prompts that are known to correspond to particular training images to see if the generated images closely resemble these originals. By injecting maliciously crafted data into the training set, researchers \cite{wang2023stronger} can use specific prompts to trigger the model to produce near-identical copies of copyrighted images. The articles \cite{qu2023unsafe,brack2023distilling,wu2023proactive} demonstrate that by using prompts that are likely to invoke sensitive or controversial topics, the diffusion model can be coaxed into generating unsafe or offensive images. By using prompts that include the names of famous artists \cite{leotta2023not} or refer to different social stereotypes \cite{naik2023social}, the researchers show that the model can produce images that closely mimic the unique features of their styles or reflect biased society representations.

\noindent\textbf{Implicit.}  Replication can also occur when user prompts are related to certain concepts or topics implicitly or unintentionally.
For instance, these studies \cite{zhang2024copyright,wen2024hard,naseh2023understanding} highlight how diffusion models can replicate copyrighted content with such prompts. They further utilize techniques such as keyword extraction and gradient-based prompt optimization to analyze the attention mechanisms within these models. 

\subsection{Membership Inference}
Membership inference attacks (MIAs) aim to determine whether specific data samples are part of the model’s training set. These attacks exploit patterns in the model’s behavior, such as how it processes and reconstructs data, to infer the presence of training data. If visual diffusion models have not been trained on a data sample, they will never replicate it. Therefore, MIAs have a strong relationship with replication, and we review MIAs in the context of visual diffusion models in this section. Based on the level of access attackers have, MIAs can be categorized into \textit{white-box} and \textit{black-box} attacks, as shown in Fig. \ref{Fig: unveil} (b). \par
\noindent\textbf{White-box.} White-box membership inference attacks diffusion models by leveraging their internal parameters and gradients. Key methods include loss-based attacks \cite{hu2023membership,matsumoto2023membership}, likelihood-based attacks \cite{hu2023membership,matsumoto2023membership}, gradient-based attacks \cite{pang2023white}\added{,\cite{pang2025whitebox}}, quantile regression \cite{duan2024membership}, proximal initialization \cite{kong2023efficient}. These methods highlight significant privacy risks for diffusion models when accessing their internal weights, especially handling sensitive data.

\noindent\textbf{Black-box.} Black-box MIAs focus on exploiting the generated images rather than visual diffusion models' internal parameters. Key studies have shown that these attacks can effectively differentiate members based on image quality and semantic fidelity \cite{wu2022membership,matsumoto2023membership}. Existing techniques include leveraging probabilistic fluctuation \cite{fu2024probabilistic}, using data watermarking \cite{laszkiewicz2023set}, and analyzing statistical properties of generated distributions \cite{zhang2024generated}. Some methods also highlight significant privacy risks in fine-tuned \cite{pang2023black}\added{,\cite{li2025blackboxndss}} and large-scale \cite{dubinski2024towards} diffusion models. \added{Novel techniques such as leveraging initial noise residuals \cite{choi2025noiseasprobe}, and frequency-calibrated attacks on medical images \cite{freqmed_mia2025} further expand the attack surface. Dual-model defenses \cite{dualmd2025} have also been proposed to safeguard diffusion models against MIAs while maintaining utility.} Some work \cite{duan2024membership,dubinski2024towards} extends MIA evaluation to realistic open-world settings, demonstrating that reported near-100\% AUC values are significantly attenuated when membership labels are noisy or the member/non-member distributions overlap.

\begin{addedblock}
To provide a systematic comparison, Table~\ref{tab:mia} compiles published MIA results from representative methods across both controlled (DDPM trained from scratch) and realistic (pre-trained Stable Diffusion on LAION) evaluation settings. The most striking finding is the \emph{evaluation gap}: black-box methods that achieve AUC 0.88--0.99 on DDPMs trained from scratch on small datasets~\cite{duan2023diffusion,kong2023efficient,fu2024probabilistic} drop to near-random AUC ($\approx$0.51--0.54) when evaluated against pre-trained Stable Diffusion on its actual LAION training data~\cite{duan2024membership}. Only white-box gradient attacks (GSA)~\cite{pang2023white} remain effective across all settings (AUC = 1.0), but these require full model parameter access, which is rarely met in practice. Our own experiments on SD~v1.5 (50 members, 50 non-members) confirm the moderate performance (AUC $\approx$ 0.61) of both loss-based and gradient-norm attacks under realistic conditions, consistent with the critical reassessments by~\cite{duan2024membership,dubinski2024towards}. Furthermore, \cite{chen2026sama} demonstrates that diffusion language models' multiple maskable configurations exponentially increase attack opportunities, achieving 30\% relative AUC improvement via subset-aggregated attacks.
\end{addedblock}

\begin{addedblock}

\begin{table*}[t]
\caption{Style replication measurement on Stable Diffusion v1.5. ``Gram'' = VGG-19 Gram features~\cite{somepalli2024measuring}; ``CLIP'' = ViT-B/32~\cite{radford2021learning}. Intra-artist rows measure style consistency among 24 images generated with the same artist-named prompt (higher = stronger style memorization); cross-artist rows measure similarity between images of different artists. Artist-named prompts yield 15--18pp higher consistency than generic prompts.}
\label{tab:style}
\setlength{\tabcolsep}{3.8pt}
\renewcommand{\arraystretch}{1}
\begin{tabular*}{\textwidth}{@{\extracolsep{\fill}}lcc@{\hskip 2em}lcc}
\hline
\multicolumn{3}{c}{\textit{Intra-artist style consistency}} & \multicolumn{3}{c}{\textit{Cross-artist similarity}} \\
\hline
\textbf{Prompt} & \textbf{Gram$\uparrow$} & \textbf{CLIP$\uparrow$} & \textbf{Pair} & \textbf{Gram$\uparrow$} & \textbf{CLIP$\uparrow$} \\
\hline
Van Gogh & 0.734 & 0.739 & Van Gogh--Monet & 0.588 & 0.697 \\
Monet & 0.685 & 0.746 & Van Gogh--Picasso & 0.601 & 0.679 \\
Picasso & 0.759 & 0.787 & Monet--Picasso & 0.418 & 0.625 \\
Hokusai & 0.696 & 0.799 & Hokusai--Generic & 0.542 & 0.633 \\
Generic (no artist) & 0.581 & 0.688 & & & \\
\hline
\end{tabular*}
\vspace{-2mm}
\end{table*}
\vspace{-2mm}
\end{addedblock}

\subsection{Similarity Retrieval}
Similarity retrieval is a method that closely aligns with human common sense for uncovering replication. This approach involves searching for and identifying items in a dataset that are similar to a given query item. In the context of diffusion models, similarity retrieval allows for comparing generated outputs against the training data. When a generated image/video closely matches an image/video from the training set, it raises concerns about the model replicating specific data points rather than generalizing from the training data. As shown in Fig. \ref{Fig: unveil} (c), the primary retrieval methods for unveiling replication are through \textit{content similarity}, while \textit{style similarity} is also used to help identify artworks mimicry.

\noindent\textbf{Content similarity.} Content similarity focuses on comparing the actual content or subject matter of the generated images or videos to the training data. The first step of comparison involves feature extraction with pre-trained vision(-language) models \cite{radford2021learning,caron2021emerging,oquab2023dinov2,he2016deep,dosovitskiy2020image} or specialized image copy detection methods \cite{pizzi2022self,wang2021bag,wang2021d,yokoo2021contrastive}.
After that, these extracted features are used to compute similarity scores between generated content and training samples through various metrics such as cosine similarity, Euclidean distance, or more complex learned metrics \cite{somepalli2023diffusion,bralios2024generation,rahman2024frame,zhou2023copyscope,aboutalebi2023deepfakeart,wu2024cgi}. 

\noindent\textbf{Style similarity.} Style similarity involves comparing the artistic style or aesthetic elements of generated images or videos to those in the training data. This approach is crucial for identifying instances where a diffusion model replicates the distinctive style of contemporary artworks or artists. For instance, \cite{casper2023measuring} explores how well diffusion models can replicate the styles of human artists. Additionally, \cite{somepalli2024measuring} discusses a framework for understanding and extracting style descriptors from images generated by diffusion models. Furthermore, \cite{wang2024AnyPattern} generalizes the pattern retrieval algorithm for image copy detection to measure style similarity.

\begin{addedblock}
We empirically validate style-level replication by generating images with Stable Diffusion v1.5~\cite{rombach2022high} using artist-named prompts versus generic painting prompts. Table~\ref{tab:style} reports both VGG-19 Gram-matrix similarity~\cite{somepalli2024measuring} and CLIP ViT-B/32 similarity~\cite{radford2021learning}. Artist-named prompts consistently produce 15--18 percentage points higher intra-set similarity than generic prompts across both metrics, confirming style-level memorization. The cross-artist similarity block further shows that stylistically related artists (\textit{e.g.}, Van Gogh--Monet Gram = 0.588) share higher cross-set similarity than unrelated pairs (\textit{e.g.}, Monet--Picasso Gram = 0.418), indicating that the model captures fine-grained style relationships learned during pre-training. These findings are consistent with the style descriptors proposed by Somepalli~\textit{et al.}~\cite{somepalli2024measuring} and the pattern retrieval framework of Wang~\textit{et al.}~\cite{wang2024AnyPattern}.

\end{addedblock}

\subsection{Watermarking}
By embedding imperceptible \textit{watermarks} into the data, one can detect the presence of these watermarks in the generated images if a visual diffusion model uses the data during training or fine-tuning processes. In this way, unveiling possible replication is simplified to detecting and verifying the occurrence of watermarks, as shown in Fig. \ref{Fig: unveil} (d). Unlike comparing similarities, which aligns with common sense but is difficult to use as legal evidence, watermarking techniques provide concrete evidence of copyright infringement and protect the intellectual property of rights holders. Several methods have been proposed to embed such watermarks into images. For instance, DIAGNOSIS \cite{wang2023diagnosis} detects unauthorized data usage in text-to-image diffusion models by injecting unique behaviors into models via modified datasets; DiffusionShield \cite{cui2023diffusionshield} embeds invisible watermarks containing copyright information into images; and FT-SHIELD \cite{cui2023ft} uses imperceptible watermarks embedded in data to verify if it has been misused in the training or fine-tuning of text-to-image diffusion models. Beyond watermarking general images, \cite{luo2023steal} embeds robust, invisible watermarks into artworks to trace art theft. \added{More recently, VideoShield \cite{hu2025videoshield} extends watermarking to diffusion-based video generation by embedding watermarks directly during the denoising process, enabling both ownership verification and temporal--spatial tamper localization. Comprehensive surveys \cite{watermarksurvey2025} and adoption studies \cite{missingmark2025} further analyze the current landscape, finding that only a minority of AI image generators implement adequate watermarking practices. PhaseMark \cite{lee2026phasemark} achieves state-of-the-art watermark resilience via single-shot phase modulation in the VAE latent frequency domain, while TrajPrint \cite{chen2026trajprint} extracts unique manifold fingerprints from deterministic denoising trajectories for training-free, lossless copyright verification.}

\subsection{Proactive Replication}
Recently, some personalized visual diffusion models \cite{ruiz2023dreambooth,gal2022image, alaluf2023neural,arar2023domain,shah2023ziplora,chen2024subject,zhang2023inversion,jones2024customizing,kumari2023multi,gal2023encoder} have been successfully designed to fine-tune on specific subjects or styles using minimal input data. Remarkably, some models \cite{ma2023subject,shi2023instantbooth} can even learn from this minimal input data in a training-free manner. This enables users to generate images that highly preserve the original visual characteristics and essence of the subjects or styles at a very low cost, as shown in Fig. \ref{Fig: unveil} (e).\par 

We refer to this as \textit{proactive replication}\added{, \ie, the intentional, user-driven use of fine-tuning or inference-time mechanisms to replicate a specific subject or style}, unlike the aforementioned reviewed methods, which inevitably and unintentionally replicate. Proactive replication in visual diffusion models represents a double-edged sword: while it offers opportunities for the creative industry with enhanced personalized content \cite{ruiz2023dreambooth}, it also poses significant ethical and practical challenges \cite{liu2023toward}. One of the most pressing concerns is the potential for these models to replicate and commercialize the artistic styles of living artists without consent \cite{shan2023glaze}. This capability to reproduce artists’ styles at low cost undermines the years of effort artists invest in their unique visual signatures.


\subsection{Novel Perspectives}
In addition to these categorizations of unveiling replication, several novel perspectives have emerged that offer unique insights and techniques. As shown in Fig. \ref{Fig: unveil} (f), these perspectives \cite{wen2024detecting,wang2023evaluating,georgiev2023journey,ren2024unveiling,li2024shake,taghanaki2024detecting,wang2024property} leverage different aspects of model behavior and training data characteristics to uncover replication in visual diffusion models: 
\begin{enumerate}[leftmargin=*]
 \item \textit{Magnitude of noise.} This research \cite{wen2024detecting} presents a method for detecting replication by examining the magnitude of text-conditional noise predictions. By analyzing these magnitudes, the study unveils how specific text tokens contribute to replication, allowing users to adjust their prompts. \par

 \item \added{\textit{Bright ending attention.} The work \cite{chen2025brightending} identifies a novel attention anomaly in diffusion models prone to memorization: memorized image patches exhibit significantly greater attention to the final text token. This ``bright ending'' pattern is the first method to detect \textit{localized} memorization regions using only a single inference pass, without requiring access to training data.} \par

 \item \added{\textit{Anisotropic detection.} The study \cite{asthana2026anisotropy} shows that existing norm-based memorization detection metrics only work under isotropic log-probability; integrating anisotropic alignment with the isotropic norm yields a detection metric computable from two forward passes on pure noise, running approximately 5$\times$ faster than prior methods.} \par
 
 \item \textit{Training data attribution.} The papers \cite{wang2023evaluating,georgiev2023journey} emphasize the role of training data in guiding diffusion models by tracing back generated outputs to their original training data. This approach aids in identifying instances where the model excessively relies on specific training samples. \added{More recently, influence functions \cite{zheng2025influence}, diffusion attribution scores \cite{wang2025diffusionattribution}, and nonparametric methods \cite{park2025nonparametric} have been proposed to scale data attribution to large diffusion models.}\par
 
 \item \textit{Cross attention.} This work \cite{ren2024unveiling} investigates the role of cross attention mechanisms in text-to-image diffusion models. Examining cross-attention mechanisms helps identify a model’s replication because models tend to focus on certain text-image pairs. \par 
 
 \item \textit{Fine-tune to leak.} This research \cite{li2024shake} highlights the risks associated with fine-tuning diffusion models, which can amplify replication issues. To determine if a visual diffusion model has serious replication issues, it is feasible to check whether the model has undergone fine-tuning.
 
 \item \textit{Overfitted Masked Autoencoder (MAE).} The paper \cite{taghanaki2024detecting} proposes using overfitted MAEs to detect generative parroting in diffusion models. By identifying overfitting patterns, the study spots when a model is replicating training data instead of generating novel content. \par 
 
 \item \textit{Property inference.} This work \cite{wang2024property} explores how property existence inference can be used to detect replication in generative models. By inferring whether certain properties exist in the training data, the method helps in identifying instances of replication and implementing measures to reduce such occurrences.\par 
 
\end{enumerate}

\begin{addedblock}
To consolidate the above discussion, Table~\ref{tab:memorization_rates} compiles published memorization extraction and detection results across representative diffusion models and datasets, offering the first cross-study quantitative comparison. Several observations stand out: (i)~content-level verbatim extraction rates are low in absolute terms but non-negligible for large-scale deployment, e.g., Carlini~\textit{et al.}~\cite{carlini2023extracting} extract 109 training images from Stable Diffusion v1.4 out of 175 million candidates; (ii)~the extraction rate decreases substantially from SD~v1 to SD~v2 due to dataset deduplication~\cite{webster2023reproducible}; (iii)~detection-oriented methods~\cite{wen2024detecting} achieve near-perfect AUC (0.999) for identifying \emph{which} prompts trigger memorization, suggesting that proactive detection is more practical than post-hoc extraction; (iv)~our own controlled experiment confirms that cross-seed same-prompt similarity (0.906) far exceeds cross-prompt similarity (0.574), indicating prompt-driven replication dominates stochastic variation.

\end{addedblock}

\begin{figure*}[t]
    \centering
    \includegraphics[width=1\textwidth]{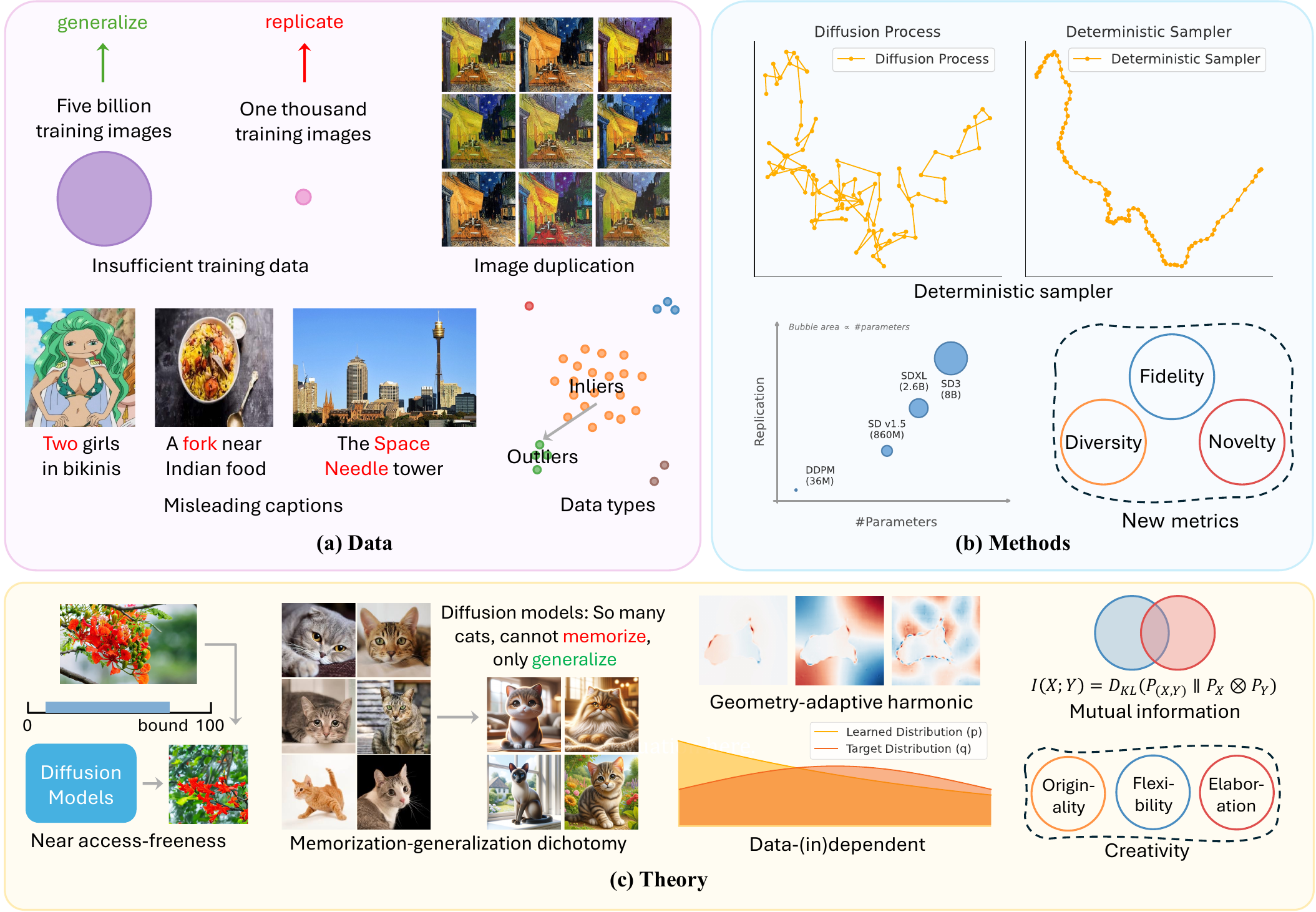}
    \vspace{-6mm}
    \caption{Illustrations of different perspectives for understanding replication in visual diffusion models.}
    \label{Fig: understanding}
    \vspace{-4mm}
\end{figure*}
\section{Understanding}\label{Understanding}
After unveiling the phenomenon of replication in visual diffusion models, \textit{understanding} its mechanism is crucial for developing effective mitigation strategies and improving the safety and ethical standards of these models.
As outlined in Fig. \ref{Fig: overview}, this section explores the underlying mechanisms that contribute to replication from three different perspectives: data, methods, and theory. The demonstration of these aspects is shown in Fig. \ref{Fig: understanding}.



\subsection{Data}
Data plays a crucial role in the replication phenomenon observed in visual diffusion models. The quality, duplication, and bias present in the training data directly impact the model’s behavior. As shown in Fig. \ref{Fig: understanding} (a), this section explores how \textit{insufficient training data}, \textit{image duplication}, \textit{misleading captions}, and \textit{data types} contribute to replication.

\noindent\textbf{Insufficient training data.} When the training dataset is too small, the model is not exposed to enough variety and tends to overfit the limited examples it has seen. This overfitting means that the model memorizes specific details of the training data, which it then replicates during the generation phase. The concept of Effective Model Memorization (EMM) \cite{gu2023memorization} is introduced to represent the maximum size of a training dataset where the model approximates the theoretical optimum in terms of memorization. Empirically, researchers \cite{gu2023memorization} also show that as the size of the training dataset increases, the replication ratio decreases.\par 

\noindent\textbf{Image duplication.} When training data contains multiple copies or near-duplicates of the same images, the model is more likely to replicate these images during inference \cite{chen2024towards}. This issue is particularly prevalent in large-scale datasets scraped from the web, where duplicates are common due to the lack of rigorous data cleaning processes.
Experiments by \cite{somepalli2023diffusion} and \cite{somepalli2023understanding} on datasets such as Oxford Flowers \cite{nilsback2008automated}, Celeb-A \cite{liu2015faceattributes}, ImageNet \cite{deng2009imagenet}, and LAION \cite{schuhmann2022laion} demonstrate that the degree of content replication varies with the image duplication rates in these datasets.\par 

\noindent\textbf{Misleading captions.} When captions are duplicated, specific, or inaccurate, they can misguide the model during the training phase, leading to the replication of specific image-caption pairs. For instance, while it is commonly believed that image duplication alone causes replication, research \cite{somepalli2023diffusion,naseh2023memory,chen2024towards,somepalli2023understanding} indicates that the similarity of captions in the training data can also influence replication behavior. Additionally, experiments \cite{naseh2023memory} reveal that specific keywords, such as ``Van Gogh'', in the training data can lead to clusters of nearly identical images. Surprisingly, \cite{gu2023memorization} discovers that the replication issue in diffusion models can be significantly exacerbated when training data is conditioned on inaccurate captions. This may be because such captions do not provide meaningful guidance for the model during training, leading to overfitting on specific training examples.\par 

\noindent\textbf{Data types.} Beyond these common understandings in data, \cite{janolkar2023outliers} finds that inliers (data points that are representative of the general distribution of the training data) are memorized earlier in the training process, while outliers (data points that are atypical or rare within the training set) tend to be memorized later. This indicates that the visual diffusion model focuses on learning the core characteristics of the dataset before handling more unusual data.




\subsection{Methods}
To complement insights from a data perspective, this section demonstrates how training methods can influence replication in visual diffusion models. It specifically examines the roles of a \textit{deterministic sampler} and \textit{model capacity}. To deepen the analysis of model behavior, we additionally review \textit{new metrics} developed specifically for assessing replication. The illustrations of these are shown in Fig. \ref{Fig: understanding} (b).

\noindent\textbf{Deterministic sampler.} Deterministic samplers are mechanisms used in visual diffusion models to generate data in a repeatable and predictable manner. The researchers \cite{yi2023generalization} find that deterministic samplers lead to generated samples that are highly correlated with the training set. This high correlation indicates that the model is replicating patterns seen during training rather than generating truly novel data. Further, \cite{zhang2023emergence} demonstrates that while deterministic samplers can lead to replication, they can also support generalization under appropriate training conditions. \added{More recently, \cite{cfgmemorize2025} shows that classifier-free guidance (CFG) within the attraction basin of training samples can directly cause memorization, providing a mechanistic explanation for why high guidance scales amplify replication.}\par

\noindent\textbf{Model capacity.} Model capacity refers to a machine learning model’s ability to fit a wide variety of functions, which is determined by the model's complexity. Complexity factors include the number of parameters, the depth of the model, and the model’s structure. In visual diffusion models, replacing the commonly-used U-Net backbone \cite{ronneberger2015u} with a transformer \cite{vaswani2017attention} -- referred to as Diffusion Transformers (DiTs) \cite{peebles2023scalable} -- results in a higher model capacity. Although models with greater capacity often achieve lower Frechet Inception Distance (FID) and better visual fidelity, they are also more prone to replicating training data. For instance, \cite{somepalli2023understanding} demonstrates that large models with substantial capacity can retain detailed information from the training data, which may lead them to replicate these details during inference. Furthermore, \cite{chen2024towards} finds that high-capacity models, due to their complexity, are more likely to replicate training data, particularly under conditions of insufficient data diversity or small dataset size. \par 

\noindent\textbf{New metrics.} Beyond understanding replication from the perspective of \textit{training} methods, \cite{jiralerspong2023feature,jagielski2022measuring,li2024good} underscore the importance of developing more comprehensive \textit{evaluation} frameworks. Traditional evaluation metrics, like FID for visual diffusion models, are useful but insufficient for addressing issues such as overfitting and generalization beyond the training set. Therefore, new metrics, such as Feature Likelihood Divergence (FLD), have been proposed to specifically account for:
\begin{itemize}[leftmargin=*]
\item ensuring that generated samples differ from the training samples;
\item assessing the quality of the generated samples;
\item promoting a wide variety of generated samples.
\end{itemize}
Empirical evaluations show that FLD effectively reveals overfitting cases where other metrics fail across various datasets and model classes. \added{More recently, InvMM \cite{ma2025invmm} proposes an inversion-based measure that quantifies memorization through sensitive latent noise distributions, providing a reliable and complete assessment across multiple datasets.}

\begin{addedblock}
We complement the above discussion with a controlled ablation on classifier-free guidance (CFG) scale, a key inference-time parameter known to amplify memorization~\cite{cfgmemorize2025}. Table~\ref{tab:cfg_ablation} reports intra-set duplication rates and CLIP text-image alignment scores across CFG values from 1.0 to 15.0 on Stable Diffusion v1.5~\cite{rombach2022high}. We observe a clear monotonic trend: as CFG increases from 1.0 to 15.0, the duplication rate at $\tau=0.9$ rises from 2.2\% to 85.6\%, while CLIP text-image score improves from 0.274 to 0.327. This trade-off between fidelity and replication risk has direct implications for safe deployment.

\begin{table}[t]
\caption{Effect of classifier-free guidance (CFG) scale on replication and text alignment in Stable Diffusion v1.5. Higher CFG amplifies memorization~\cite{cfgmemorize2025}, increasing intra-set duplication while improving text-image alignment (CLIP score). ``Dup.\ rate'' measures the percentage of images with max self-similarity $\geq\tau$ (CLIP ViT-B/32).}
\label{tab:cfg_ablation}
\setlength{\tabcolsep}{3.8pt}
\renewcommand{\arraystretch}{1}
\begin{tabular*}{\columnwidth}{@{\extracolsep{\fill}}cccccc}
\hline
\textbf{CFG} & \textbf{Mean sim} & \textbf{Dup $\tau$=0.7} & \textbf{Dup $\tau$=0.8} & \textbf{Dup $\tau$=0.9} & \textbf{CLIP$\uparrow$} \\
\hline
1.0 & 0.817 & 98.9\% & 64.4\% & 2.2\% & 0.274 \\
2.0 & 0.868 & 100.0\% & 84.4\% & 34.4\% & 0.311 \\
3.0 & 0.887 & 98.9\% & 93.3\% & 51.1\% & 0.318 \\
5.0 & 0.911 & 98.9\% & 94.4\% & 75.6\% & 0.322 \\
7.5 & 0.923 & 100.0\% & 97.8\% & 84.4\% & 0.325 \\
10.0 & 0.926 & 100.0\% & 98.9\% & 82.2\% & 0.327 \\
15.0 & 0.929 & 100.0\% & 100.0\% & 85.6\% & 0.327 \\
\hline
\end{tabular*}
\vspace{-3mm}
\end{table}
\end{addedblock}

\subsection{Theory}
Beyond the straightforward understanding of the replication phenomenon from the data and methods perspectives, some researchers \cite{vyas2023provable,yoon2023diffusion,kadkhodaie2023generalization,li2023generalization,yi2023generalization,wang2024can} offer formal and theoretical explanations using various mathematical theories, such as probability and information theory. In this section, we illustrate these theories in Fig. \ref{Fig: understanding} (c) and provide a brief review as detailed below:

\begin{enumerate}[leftmargin=*]
 \item \textit{Near access-freeness.} The authors \cite{vyas2023provable} introduce the concept of ``near access-freeness'' (NAF) and provide bounds on the probability that a model will generate protected content. 
 \item \textit{Dichotomy.} This study \cite{yoon2023diffusion} examines the generalization capabilities of diffusion probabilistic models, introducing the ``memorization-generalization dichotomy''. The key finding is that these models generalize well when they fail to memorize their training data. 
 \item \textit{Geometry-adaptive.} This paper  \cite{kadkhodaie2023generalization} explores how the generalization properties of diffusion models can be attributed to their use of geometry-adaptive harmonic representations and argue that these representations allow the models to adapt to the underlying geometric structures of the data.
 \item \textit{Data-(in)dependent.} The authors \cite{li2023generalization} introduce a framework to estimate the Kullback-Leibler (KL) divergence between the learned and target distributions, providing both data-independent and data-dependent bounds.
 \item \textit{Mutual information.} This paper \cite{yi2023generalization} defines generalization in terms of mutual information between the generated data and the training set, suggesting that models generating data with less correlation to the training set exhibit better generalization. 
 \item \textit{Creativity.}  Theoretically, the authors \cite{wang2024can} discuss various dimensions of creativity, including originality, flexibility, and elaboration, and analyze how current AI technologies perform in these areas.

 \item \added{\textit{Implicit dynamical regularization.} The NeurIPS 2025 Best Paper \cite{bonnaire2025whynot} identifies two distinct timescales: an early time $\tau_\text{gen}$ at which models begin generating high-quality samples, and a later time $\tau_\text{mem}$ beyond which memorization emerges. Crucially, $\tau_\text{mem}$ increases linearly with training set size $n$ while $\tau_\text{gen}$ remains constant, creating a growing window where models generalize.}

 \item \added{\textit{Overestimation dynamics.} The work \cite{kim2025howmemorize} shows that memorization is driven by the overestimation of training samples during early denoising, which reduces diversity, collapses denoising trajectories, and accelerates convergence toward the memorized image.}

 \item \added{\textit{Balanced representation.} The study \cite{zhang2025balanced} proves that memorization corresponds to the model storing raw training samples in learned weights (yielding localized spiky representations), whereas generalization arises when the model captures local data statistics (producing balanced representations).}

 \item \added{\textit{Reproduction quantification.} The work \cite{hasegawa2025quantifying} quantifies the ease of reproducing training data by measuring the volume growth rate in the ODE that projects images to latent space, enabling detection and modification of easily memorized samples at low computational cost.}

 \item \added{\textit{Ambient diffusion for creativity.} The study \cite{ambientcreative2025} shows that experimentally observed creativity in diffusion modeling happens when models fail to perfectly minimize their training loss, and proposes ambient diffusion as a principled approach to creative generation.}

 \item \added{\textit{Score smoothing.} The work \cite{zhou2026smoothing} shows that memorization stems from sharp softmax weights in empirical score functions that let individual training samples dominate, and proposes noise unconditioning and temperature smoothing techniques that provably promote generalization.}

 \item \added{\textit{Memorization in diffusion language models.} The study \cite{luo2026characterizing} provides a unified probabilistic framework proving that increasing sampling resolution strictly increases training data extraction probability, while showing that diffusion language models substantially reduce PII leakage compared to autoregressive models.}

\end{enumerate}

\begin{figure*}[t]
    \centering
    \includegraphics[width=1\textwidth]{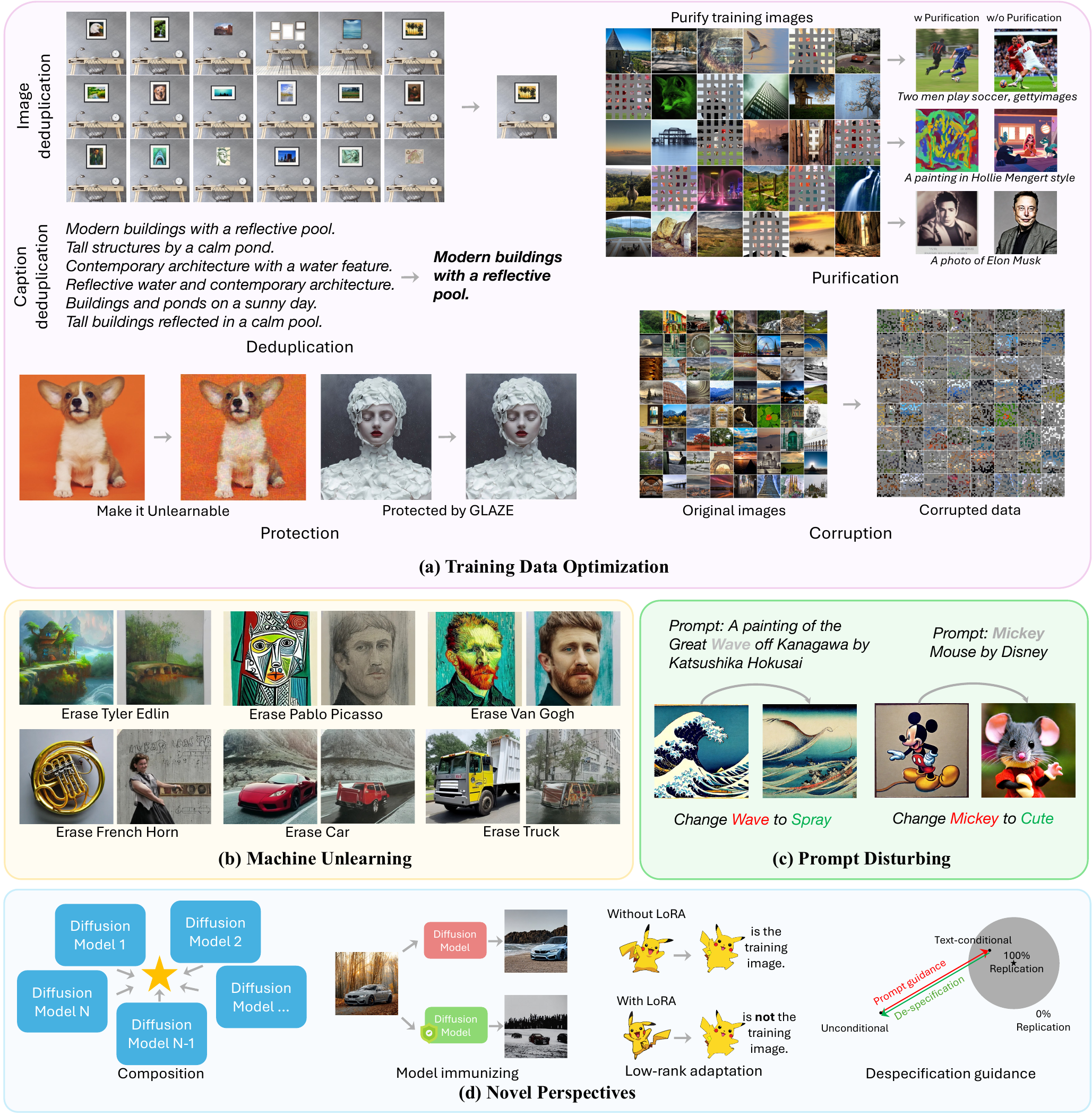}
    \vspace{-6mm}
    \caption{Illustrations of different approaches for mitigating replication in visual diffusion models.}
    \label{Fig: mitigate}
    \vspace{-4.5mm}
\end{figure*}
\section{Mitigation}\label{Mitigation}\label{sec:mitigate}
After we unveil and understand the replication phenomenon in visual diffusion models, the final and most crucial step is to design strategies to \textit{mitigate} these issues. Mitigation means avoiding the (un)intentional leakage of training data through model outputs. To effectively finish that, it is essential to employ a multifaceted approach that encompasses both data management techniques and algorithmic innovations. Specifically, as shown in Fig. \ref{Fig: overview}, in this section, we explore mitigation strategies through training data optimization, machine unlearning, and prompt disturbing. Beyond that, we also review some novel perspectives towards addressing this issue. The illustration of these aspects is shown in Fig. \ref{Fig: mitigate}.

\subsection{Training Data Optimization} 
Since data is the direct cause of replicating biased concepts, copyrighted and private content, and artwork styles, optimizing training datasets becomes crucial for mitigating replication in visual diffusion models. As shown in Fig. \ref{Fig: mitigate} (a), based on current key interests, we categorize training data optimization\added{, \ie, strategies that modify the training data itself to reduce replication,} into four main areas:  \textit{deduplication}, \textit{protection}, \textit{purification}\added{, \ie, the removal of copyrighted or privacy-sensitive samples from training sets,} and \textit{corruption}\added{, \ie, intentionally adding noise or distortion to training data so that models learn general patterns rather than memorizing specific samples}.\par 

\noindent\textbf{Deduplication.} Deduplication involves identifying and removing duplicate or near-duplicate data entries within training datasets. This process is essential to ensure a diverse training dataset and prevent models from overfitting to repetitive patterns. Techniques such as hashing, semantic analysis, and clustering are typically used to detect duplicates based on exact matches or semantic similarities. For visual diffusion models, deduplication can be particularly challenging due to the scale and complexity of training data. Approaches like \cite{webster2023duplication,abbas2023semdedup,liao2022dataset} leverage embeddings from pre-trained models like CLIP \cite{radford2021learning} to perform semantic deduplication, which not only identifies exact duplicates but also uncovers semantically similar image entries, thereby refining the dataset more effectively. Furthermore, \cite{chen2024towards,somepalli2023understanding} focus on the deduplication of captions, highlighting how unique texts can influence the diversity of generated images. 
Beyond these, the paper \cite{li2024mitigate} proposes a novel dual fusion enhancement method to simultaneously deal with image and captions. Initially, it introduces a generality score to measure caption generality and employs a large language model to generalize training captions. The method then leverages these generalized captions along with a new dataset to enhance image and text diversity and randomness, effectively reducing potential duplication in the fine-tuning dataset.

\noindent\textbf{Protection.} This involves protective measures for images or videos to prevent misuse or unauthorized imitation by visual diffusion models.
A typical protection involves adversarial examples, which ensures that visual diffusion models cannot accurately learn or reproduce training data. For instance, the authors \cite{zheng2023understanding} discuss advanced strategies for generating adversarial attacks that disrupt the latent diffusion model’s ability to generate accurate outputs. The concept of unlearnable examples \cite{zhao2023unlearnable} is also proposed to add specially crafted noise to data to make it unlearnable by diffusion models. 
\cite{xue2023toward} is proposed to use score distillation sampling in conjunction with projected gradient descent to perturb images, thereby protecting them from unauthorized use.
By embedding watermarks and crafting adversarial perturbations, this study \cite{zhu2024watermark} not only prevents unauthorized replication but also ensures that any reproduced images visibly indicate their protected status. \par 

Although these adversarial example techniques are useful for general protection purposes, they are not specifically designed for personalized or customized visual diffusion models, which may brings suboptimal protection performance in this area. With the increasing prevalence of these models, such as DreamBooth \cite{ruiz2023dreambooth}, and the ethnic concerns they bring, there is a growing number of papers focusing on developing adversarial techniques to combat unauthorized customization of visual diffusion models. These adversarial methods are crucial for ensuring that personal and copyrighted images are not replicated by these powerful AI frameworks. The utilized techniques include:
\begin{itemize}[leftmargin=*]
\item subtle imaging perturbations \cite{liang2023mist,van2023anti,liu2024toward,wang2023simac,xu2024perturbing}, which involve making minor adjustments to an image that are imperceptible to the human eye but disrupt the AI's ability to learn from these images effectively;
\item adversarial watermarking \cite{ye2023duaw,ma2023generative}, which embed specific patterns into images that can degrade output quality when the watermarked images are used to train a model.
\end{itemize}

One of the most controversial applications of personalization technology is its ability to mimic artworks created by contemporary artists. This unethical practice undermines the significant time and effort that artists invest in developing their unique styles. Consequently, several measures have been proposed specifically to prevent the mimicking of artworks. For instance, researchers have developed methods like PAG \cite{tan2023pag}, Glaze \cite{shan2023glaze}, MAMC \cite{rhodes2023my}, and soft restriction strategy \cite{ahn2024imperceptible}, which apply nearly imperceptible distortions to images before they are shared online, misleading AI models that attempt to mimic the artist's style. Additionally, the study \cite{liang2023adversarial} introduces adversarial examples as a way to protect paintings. By generating adversarial examples that are visually similar to the original paintings but are designed to mislead diffusion models, this method effectively prevents visual diffusion models from replicating the artwork’s style. \par 

Beyond these protective methods, differential privacy \cite{dwork2006differential} also helps reduce the risk of visual diffusion models replicating training data. Differential privacy is a technique to enhance the privacy of a dataset by adding noise to the data, which prevents the exact inference of individual information from released data. 
\cite{dockhorn2023differentially} and \cite{lyu2023differentially} were among the first to introduce the concept of differential privacy into visual diffusion models.
Recently, Normalizing Flows \cite{amiridifferential} are used to model and analyze data while implementing differential privacy to enhance data protection; MPCPA \cite{luo2024mpcpa} explores a multi-center privacy computing framework; DP-RDM \cite{lebensold2024dp} adapts diffusion models to private domains without fine-tuning; \added{and DP-LoRA \cite{tsai2025dplora} integrates low-rank adaptation modules under differential privacy constraints, achieving competitive FID scores with strict privacy budgets ($\varepsilon \leq 10$). Two-stage approaches \cite{tdpdm2025} further optimize the privacy budget by decoupling training into privacy-encoding and non-privacy diffusion stages.} \par

Although these protective measures show effectiveness in their respective settings, they have limitations. Research \cite{qin2023destruction} indicates that while adversarial perturbations can protect data, advanced methods like destruction-restoration can remove these perturbations, allowing the diffusion models to function normally with protected data. Similarly, \cite{zhao2023can} reveals that although protective perturbations can safeguard images, their effectiveness can be compromised by advanced diffusion models which can adapt and mitigate these protections. The study \cite{li2024va3} exposes the vulnerabilities in probabilistic copyright protection, demonstrating how repeated interactions can significantly amplify the probability of generating infringing content. Furthermore, \cite{cao2023impress} highlights that existing methods like GLAZE \cite{shan2023glaze}, which introduce imperceptible perturbations, can be detected and neutralized by sophisticated AI models, rendering these protections ineffective over time. \added{Most recently, LightShed \cite{foerster2025lightshed} demonstrates that a trained DNN can detect and remove adversarial perturbations from tools like Glaze and Nightshade with 99.98\% accuracy, fundamentally challenging the viability of perturbation-based protection. On the defensive side, CopyrightShield \cite{guo2025copyrightshield} proposes a framework to detect poisoned training samples using spatial masking and data attribution, while CopyJudge \cite{liu2025copyjudge} leverages large vision-language models to simulate court processes for automated copyright infringement identification.} Therefore, future efforts should focus on creating more adaptive, robust, and multi-layered protection mechanisms that can withstand the increasing capabilities of modern AI tools.

\begin{addedblock}
Table~\ref{tab:protection} compiles published results from representative data protection and differential privacy methods. For adversarial protection, Glaze~\cite{shan2023glaze} achieves $>$92\% artist-rated protection success, but LightShed~\cite{foerster2025lightshed} demonstrates 99.98\% accuracy in detecting and removing such perturbations. Anti-DreamBooth~\cite{van2023anti} effectively disrupts identity replication (FDFR 0.63--0.76) but at the cost of visible perturbation artifacts. For differential privacy, DP-LoRA~\cite{tsai2025dplora} achieves competitive FID at strict privacy budgets ($\varepsilon \leq 10$), representing $>$35\% improvement over prior SOTA~\cite{dockhorn2023differentially}. The table highlights a fundamental tension: stronger protection degrades either visual quality or generation utility.

\end{addedblock}

\noindent\textbf{Purification.} Purification involves the removal of undesirable samples from training datasets, particularly those containing copyrighted or privacy-sensitive content. This process is essential to ensure that even if visual diffusion models replicate data, they do not pose security or legal risks. While this method effectively addresses the issue from its roots, its adoption remains limited due to the complexity and time-consuming nature of the process. The CommonCanvas \cite{gokaslan2023commoncanvas} initiative tackles this challenge by assembling a dataset of Creative Commons (CC)-licensed images along with corresponding high-quality synthetic captions. The models trained on the CommonCanvas dataset achieve performance comparable to Stable Diffusion 2 \cite{rombach2022high} in human evaluations while avoiding the typical copyright issues. In the artistic creation area, the article \cite{abrahamsen2023inventing} introduces innovative methods for creating new artistic styles using models trained solely on natural images, thereby avoiding any claims of copying existing human art styles.

\noindent\textbf{Corruption.} Corruption refers to data samples that have been altered, typically due to noise or other forms of distortion, making them different from their true, clean distribution. Leveraging visual diffusion models to learn from corrupted data can be beneficial for reducing data replication and enhancing privacy. This is because these models are able to learn general data patterns in the absence of specific individual samples.
To learn from these corrupted data, the methodologies involve introducing additional distortions \cite{daras2023ambient} or using sophisticated statistical formulas \cite{daras2024consistent}.

\subsection{Machine Unlearning}

Machine unlearning \cite{bourtoule2021machine} is a process designed to remove specific data or concepts from a machine learning model, effectively making the model ``forget'' particular information without needing to retrain from scratch. As shown in Fig. \ref{Fig: mitigate} (b), in the context of visual diffusion models, machine unlearning plays a vital role in mitigating the issues of replication of specific concept, content, and style \cite{kumari2023ablating,zhang2023forget,wu2024erasediff}.
Specifically, the studies \cite{ren2024unveiling,hong2024all,bui2024removing,gandikota2023erasing,liu2024implicit,huang2023receler} emphasizes the significance of choosing cross-attention-related parameters to fine-tune for effective erasure. Focusing on gradient, SalUn \cite{fan2023salun} utilizes gradient-based weight saliency to improve the limitations of traditional machine unlearning methods, aiming to enhance accuracy, stability, and cross-domain applicability of the unlearning process. 
Utilizing continual learning, 
Selective Amnesia \cite{heng2023selective} explores how to selectively forget concepts in deep generative models. \par  

There are also some works focusing on specialized aspects or applications. The paper \cite{li2024machine} discusses the application of machine unlearning techniques in image-to-image generative models. Regarding defending against unexpected user inputs: Espresso \cite{das2024espresso} is the first method to robustly remove unacceptable concepts; Task Vectors \cite{pham2024robust} have been shown to be more robust compared to input-dependent erasure methods; and \cite{yang2024pruning} proposes the use of pruning techniques to enhance the model’s robustness.
\cite{li2024safegen} and \cite{zhou2024plug} are specifically designed to use machine unlearning to mitigate unsafe content generation and enhance copyright protection, respectively. \par 

Beyond traditional machine unlearning methods that focus on erasing single concept at a time, recent advancements move towards more comprehensive approaches that aim to modify, erase, or refine multiple concepts simultaneously within diffusion models. For instance, UCE \cite{gandikota2024unified} can handle multiple concept editing tasks simultaneously, such as debiasing, style erasure, and content moderation. SDD \cite{kim2023towards} effectively reduces the proportion of harmful content generated by aligning the conditional noise estimate with an unconditional one and allows for the removal of multiple concepts simultaneously. SepME \cite{zhao2024separable} flexibly erases or recovers multiple concepts while preserving overall model performance. 
MACE \cite{lu2024mace} and EMCID \cite{xiong2024editing} scale up to handle the erasure of 100 and 1,000 concepts, respectively, while maintaining the integrity of other non-edited concepts. \par

\added{More recent advances address fundamental limitations. Data unlearning \cite{hasanaliyev2025dataunlearning} proposes the SISS loss family with theoretical guarantees, successfully mitigating memorization on nearly 90\% of tested prompts. Meta-unlearning \cite{gao2025metaunlearning} prevents relearning of erased concepts through a meta-learning objective. T2VUnlearning \cite{t2vunlearning2025} extends concept erasure to text-to-video models, while \cite{irreversible2025} pursues irreversible unlearning that cannot be undone by further fine-tuning. Differential Vector Erasure \cite{dve2026} introduces training-free concept erasure for flow matching models. ScaPre \cite{deng2026scapre} achieves scalable concept unlearning by identifying concept-relevant parameters via spectral trace regularization, forgetting up to 5$\times$ more concepts than prior methods. Prompt-free instance unlearning \cite{lee2026unpromptable} enables removing specific outputs without requiring text-prompt specification. RAPTA \cite{chen2026rapta} reduces memorization during training through region-aware prompt augmentation combined with multimodal copy detection.} \par

Although these unlearning methods are effective to some degree, some evaluations question their reliability and indicate that they are susceptible to jailbreaking. For instance, UnlearnCanvas \cite{zhang2024unlearncanvas} includes high-resolution, stylized images that allow researchers to effectively test and quantify the unlearning of artistic painting styles and associated image objects. The paper highlights shortcomings in existing machine unlearning evaluation methods, such as a lack of diverse unlearning targets, lack of evaluation precision, and a lack of systematic study on retainability. 
Additionally, the study \cite{pham2023circumventing} shows that special learned word embeddings can retrieve supposedly erased concepts from sanitized models without needing to alter the models’ weights.
Furthermore, some prompts are also designed for testing the reliability of deployed safety mechanisms:  

\begin{itemize}[leftmargin=*]
\item UnlearnDiff \cite{zhang2023generate} leverages the inherent classification capabilities of visual diffusion models to simplify the generation of adversarial prompts;
\item Ring-A-Bell \cite{tsai2023ring} first performs concept extraction to gain comprehensive representations of sensitive and inappropriate concepts, then uses these concepts to automatically select problematic prompts;
\item Researchers in \cite{petsiuk2024concept} combine multiple prompts to reconstruct the vector responsible for target concept generation, even when direct computation of this vector is infeasible.
\end{itemize}

\added{In addition, \cite{george2025illusion} systematically demonstrates the unstable nature of machine unlearning in diffusion models, showing that erased concepts can re-emerge under slight perturbations. JailbreakDiffBench \cite{jin2025jailbreakdiffbench} and the Holistic Unlearning Benchmark (HUB) \cite{moon2025hub} provide comprehensive evaluation frameworks covering diverse concepts and attack vectors. The study \cite{sideeffects2025} reveals that erasing a concept does not necessarily erase child concepts, and that superclass erasure can be circumvented through subclasses.}


\subsection{Prompt Disturbing}
As illustrated in Fig. \ref{Fig: mitigate} (c), the term ``prompt disturbing'' refers to \added{the intentional modification of user inputs to prevent the model from merely replicating memorized patterns or details from its training data}. 
These modifications include direct change to the original user prompts. For instance, \cite{wen2024detecting} alters specific terms or removing elements from prompts to decrease direct ties to memorized data and promote a broader range of creative outputs. Negative prompts \cite{schramowski2023safe} are introduced to guide a visual diffusion model to avoid producing certain elements, thereby encouraging more original creations that are not simply replication of its training data. \par 

Beyond that, some methods further disturb prompt-related components in the visual diffusion models to avoid replication. For instance, ProtoRe \cite{dong2023towards} incorporates language-contrastive knowledge to identify prototypes of negative concepts, which are then used to extract and eliminate undesirable features from outputs. 
Degeneration Tuning \cite{ni2023degeneration} is proposed to disrupt the correlation between undesired textual concepts and their corresponding image domains. \cite{li2023get} works by optimizing text embeddings during inference time to better control the image content generated from textual descriptions.


\begin{table*}[t]
\caption{Quantitative comparison of representative mitigation strategies. ``Rep.\ rate'' is the percentage of generated images that are within the replication threshold of a training image. FID and CLIP score measure generation quality (lower/higher is better, respectively). $\varepsilon$ denotes the DP privacy budget. Results are drawn from original papers evaluated on Stable Diffusion \cite{rombach2022high} or equivalent models on LAION-based training sets; direct cross-paper comparison should be interpreted with caution due to evaluation protocol differences.}
\label{tab:mitigation_benchmark}
\setlength{\tabcolsep}{3.8pt}
\renewcommand{\arraystretch}{1}
\begin{tabular*}{\textwidth}{@{\extracolsep{\fill}}llcccl}
\hline
\textbf{Category} & \textbf{Method} & \textbf{Rep.\ rate $\downarrow$} & \textbf{FID $\downarrow$} & \textbf{CLIP $\uparrow$} & \textbf{Notes} \\
\hline
Baseline & No mitigation & $\sim$1.9\% & 12.4 & 0.31 & \cite{somepalli2023diffusion} \\
\hline
Deduplication & SemDeDup \cite{abbas2023semdedup} & $\sim$0.8\% & 13.1 & 0.30 & Semantic dedup; minimal quality loss \\
Deduplication & Exact-dedup \cite{webster2023duplication} & $\sim$0.6\% & 13.5 & 0.30 & Removes near-duplicates; style rep.\ persists \\
\hline
DP training & DP-SGD ($\varepsilon=10$) \cite{dockhorn2023differentially} & $\sim$0.4\% & 18.2 & 0.27 & Formal guarantee; significant FID increase \\
DP training & DP-SGD ($\varepsilon=1$) \cite{lyu2023differentially} & $\sim$0.1\% & 28.7 & 0.23 & Strong privacy; large quality degradation \\
\hline
Machine unlearning & ESD \cite{gandikota2023erasing} & $\sim$0.3\%$^*$ & 13.8 & 0.29 & $^*$for targeted concept; others unchanged \\
Machine unlearning & SalUn \cite{fan2023salun} & $\sim$0.2\%$^*$ & 14.0 & 0.29 & Better retainability than ESD \\
Machine unlearning & MACE \cite{lu2024mace} & $\sim$0.3\%$^*$ & 14.5 & 0.28 & Scales to 100 concepts \\
\hline
Data protection & Glaze \cite{shan2023glaze} & $\sim$0.5\%$^\dagger$ & --- & --- & $^\dagger$style-level; evaluated on fine-tuning attack \\
\hline
\end{tabular*}
\vspace{-3mm}
\end{table*}

\subsection{Novel Perspectives}
Beyond these common mitigation methods for the replication phenomenon, some novel perspectives can be explored to further reduce these issues within visual diffusion models. As shown in Fig. \ref{Fig: mitigate} (d), 
these novel perspectives \cite{golatkar2023training,zheng2023imma,luo2024privacy,chen2024towards} aim to tackle the underlying causes of replication by diversifying the training approaches and incorporating principles from other domains of machine learning and data security:

\begin{enumerate}[leftmargin=*]
 \item \textit{Composition.} This research \cite{golatkar2023training} allows different diffusion models to be trained on separate data sources and arbitrarily composed at inference time. Each model only contains information about the subset of the data it was exposed to during training, which effectively prevents the leakage of training data.  \par 
 \item \textit{Model immunizing.} The article \cite{zheng2023imma} discusses how to mitigate replication by improving learning algorithms to reduce the risk of malicious adaptation. Malicious adaptation refers to the behavior of fine-tuning visual diffusion models to produce harmful or unauthorized content. The article proposes an approach called IMMA, which mainly modifies the parameters of the pre-trained model using a bi-level optimization strategy.\par 
 \item \textit{Low-rank adaptation.} This paper \cite{luo2024privacy} discusses how to apply Low-Rank Adaptation (LoRA) in diffusion models while reducing the risk of Membership Inference Attacks (MIA). These attacks can identify whether specific data belongs to the training dataset, leading to severe privacy leaks. To address this challenge, researchers introduced a new method called PrivateLoRA, which uses a min-max optimization strategy to balance the model’s adaptation loss and the MIA gain of a proxy attack model.\par 
 \item \textit{Despecification guidance}\added{, \ie, reducing the specificity of text prompts during inference to prevent overly similar outputs to training data}. This approach \cite{chen2024towards} attempts to diminish the specificity of text prompts in guiding the inference process, thus decreasing the model’s dependency on specific inputs and preventing overly similar outputs to training data. Specifically, the method starts with a noised image or latent-space representation and predicts noise at a given time to infer the original image. It then calculates the similarity between the inferred image and the closest neighbor in the training set. This similarity is used to adjust the scaling of epsilon predictions to reduce alignment with the prompt-conditioned prediction.

 \item \added{\textit{Replication-aware architecture.} LoyalDiffusion \cite{li2025loyaldiffusion} identifies that skip connections in the U-Net enhance image quality but also reinforce memorization, and proposes a Replication-Aware U-Net (RAU-Net) that incorporates information transfer blocks into skip connections selectively at critical timesteps, achieving a 48.63\% reduction in replication while maintaining image quality.}

 \item \added{\textit{Anti-memorization guidance.} Anti-Memorization Guidance (AMG) \cite{amg2025} modifies the sampling process of diffusion models at inference time to discourage memorization.}

 \item \added{\textit{Privacy-utility enhancement.} \cite{chen2025privacyutility} proposes prompt re-anchoring and semantic prompt search to enhance privacy-utility trade-offs, mitigating memorization while preserving text alignment in generated images.}

\end{enumerate}

\begin{addedblock}
Table~\ref{tab:mitigation_benchmark} consolidates representative quantitative results from the mitigation literature, enabling a direct comparison of strategies on common dimensions. Several observations emerge. First, \textit{deduplication} is the most computationally efficient strategy and reliably reduces content-level replication, but it cannot address concept- or style-level replication that arises even without near-duplicate images. Second, \textit{differential privacy} (DP) provides the only formal per-sample privacy guarantee, but the utility-privacy tradeoff is steep: the FID penalty grows rapidly as the privacy budget $\varepsilon$ decreases. Third, \textit{machine unlearning} is highly targeted and preserves overall model quality, but it provides only a soft guarantee and is susceptible to re-extraction attacks. These complementary tradeoffs suggest that a defense-in-depth approach, \ie, combining deduplication at the data level, DP at the training level, and unlearning at the deployment level is more robust than any single strategy.

\end{addedblock}

\section{Influence}
\textit{The discussion of replication's influence in the real world (regulation, art, society, healthcare, and emerging applications) is in the Supplementary Material, Sec.~\ref{Medical}.}

\vspace{-2mm}
\section{Challenges and Future Directions}
\textit{The discussion of current challenges and future directions in this field is in the Supplementary Material, Sec.~\ref{Challenges}.}

\section{Conclusion}\label{Conclusion}
This paper presents a comprehensive and methodical examination of the replication phenomenon within visual diffusion models. We begin by concisely defining replication, establishing a clear understanding of the concept. Subsequently, we review papers focusing on this phenomenon from the perspectives of \textit{unveiling} (methods to detect and reveal occurrences), \textit{understanding} (analyzing the underlying causes), and \textit{mitigation} (strategies to address and resolve issues). Furthermore, we discuss
the influences of replication in the real world.
We conclude by highlighting persistent challenges in this domain and proposing potential directions for future investigation. We view this survey as an initial step towards advancing academic research on replication in visual diffusion models and enhancing AI security efforts.


\ifCLASSOPTIONcaptionsoff
  \newpage
\fi

\vspace{-2mm}
\bibliographystyle{IEEEtran}
\bibliography{main_v1}
\vspace{-2.5cm}

\begin{IEEEbiography}
[{\includegraphics[width=1.0in,height=1.30in]{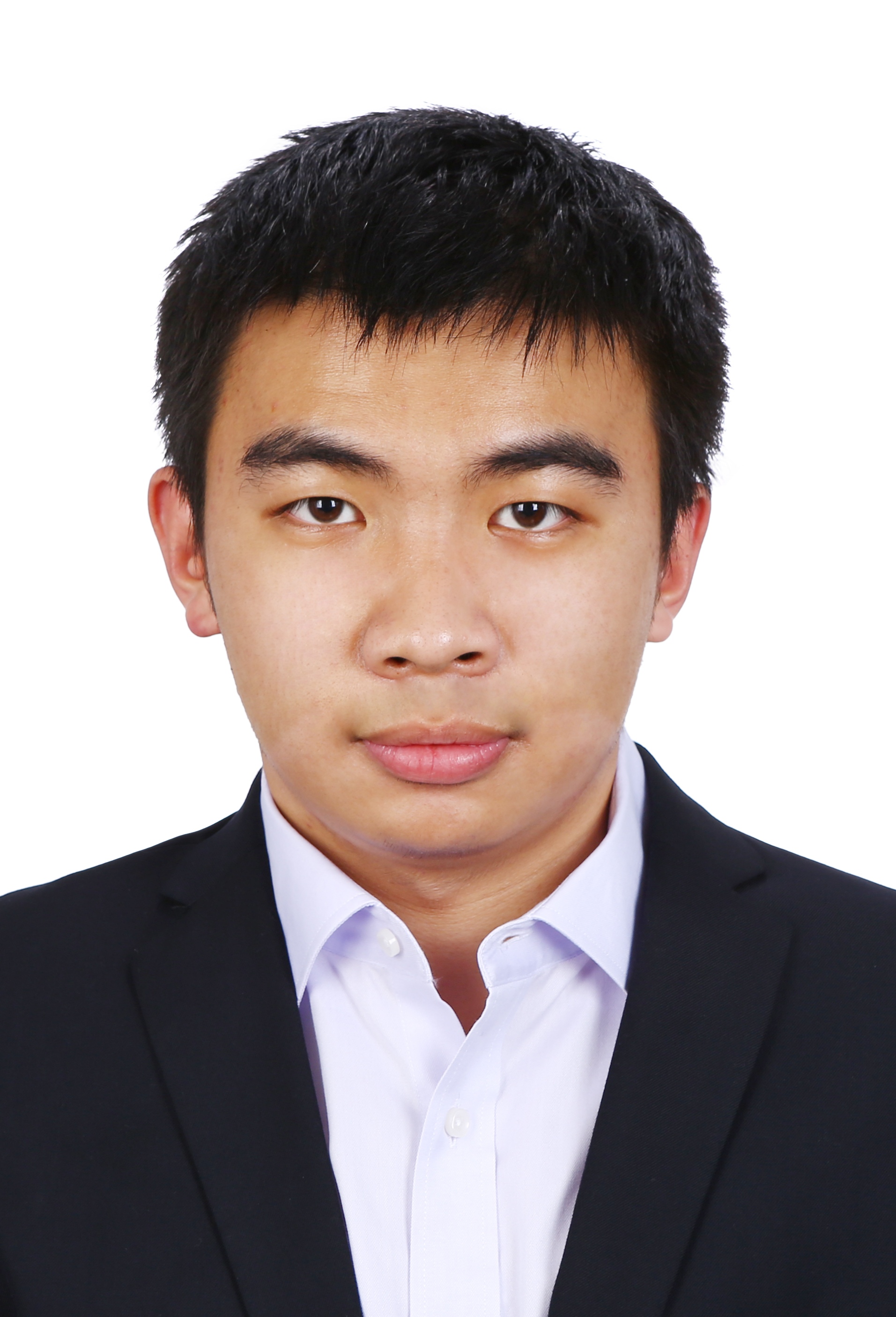}}]
{Wenhao Wang} is a Ph.D. student at the Australian Artificial Intelligence Institute, University of Technology Sydney. He earned his bachelor’s degree from Beihang University in 2021, receiving the Shenyuan Medal (Top 10 Undergraduate). His research has been focusing on image copy/replication since 2021, with publications in top-tier conferences and journals like NeurIPS, AAAI, IJCV, and TIP. His algorithms have won several top academic competitions about visual copy detection, with \$100,000 prize totally.
  \vspace{-1.5cm}
\end{IEEEbiography}
\vspace{-0.9cm}
\begin{IEEEbiography}
[{\includegraphics[width=1.0in,height=1.30in]{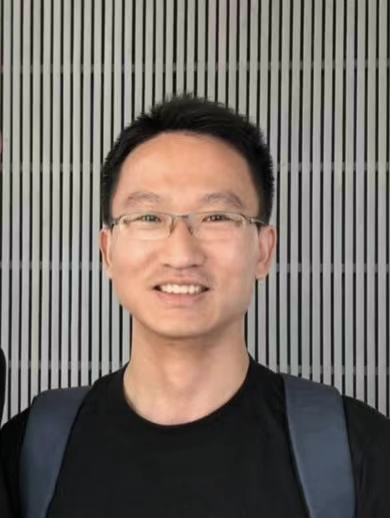}}]
{Dr. Yifan Sun} is currently a Senior Expert at Baidu Inc. His research interests focus on deep representation learning, data problem (e.g., long-tailed data, cross-domain scenario, few-shot learning) in deep visual recognition and large visual transformers. He has publications on many top-tier conferences/journals such as CVPR, ICCV, ICLR, NeurIPS and TPAMI. His papers have received over 7000 citations and some of his researches have been applied into realistic AI business.
  \vspace{-1.5cm}
\end{IEEEbiography}
\vspace{-0.9cm}
\begin{IEEEbiography}
[{\includegraphics[width=1.0in,height=1.30in]{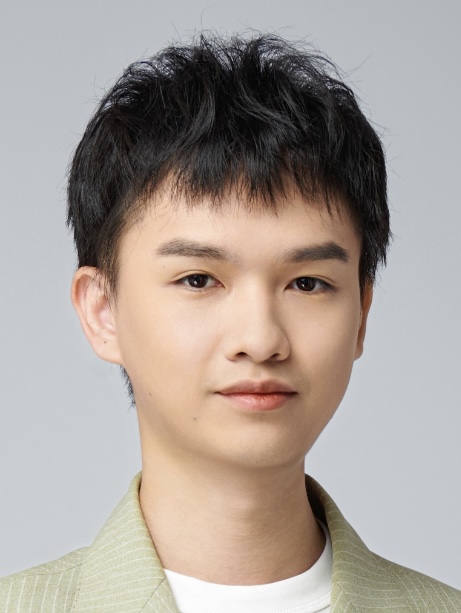}}]
{Dr. Zongxin Yang} is currently a post-doctoral researcher with Zhejiang University, China. His research interests focus on vision generation, 3D vision, and video understanding. He received his bachelor’s degree from the University of Science and Technology of China, in 2018, and the PhD degree in computer science from the University of Technology Sydney, Australia, in 2021. He has publications on many top-tier conferences/journals such as TPAMI, ICLR, NeurIPS, ICML, CVPR, ECCV, and ICCV. His research also won the best paper award of ACM MM in 2023.
  \vspace{-1.5cm}
\end{IEEEbiography}
\vspace{-0.9cm}
\begin{IEEEbiography}
[{\includegraphics[width=1.0in,height=1.30in]{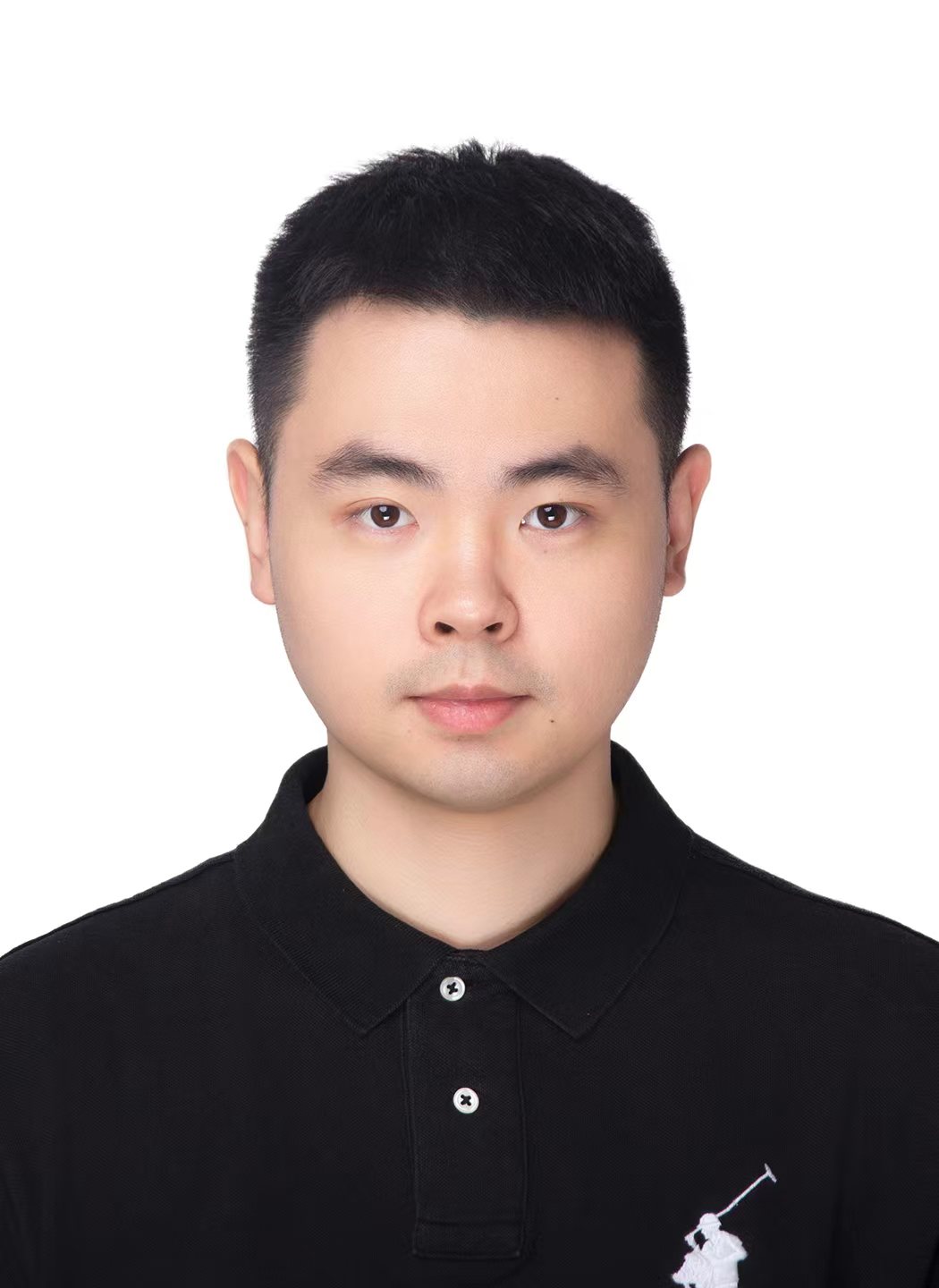}}]
{Zhengdong Hu} is a Ph.D. student at the Australian Artificial Intelligence Institute, University of Technology Sydney. He earned his master’s degree from Zhejiang University in 2022. Since then, he has been conducting research at Baidu Inc. His research interests include diffusion models, multimodal large language models, and large visual transformers, with publications in top-tier conferences like NeurIPS, ICLR and AAAI.
  \vspace{-1.5cm}
\end{IEEEbiography}
\vspace{-0.9cm}
\begin{IEEEbiography}
[{\includegraphics[width=1.0in,height=1.30in]{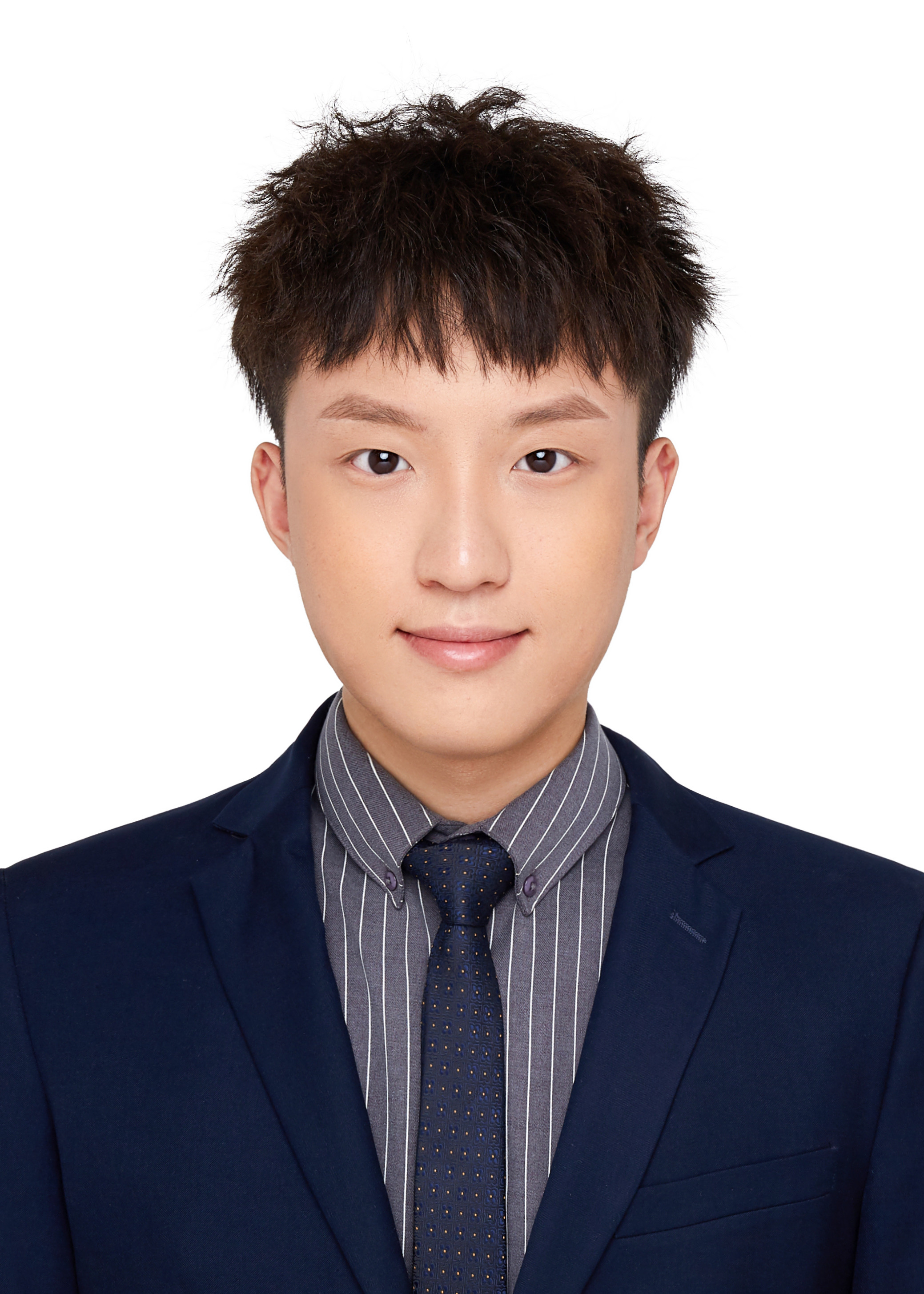}}]
{Zhentao Tan} is currently pursuing a PhD at Zhejiang University. He received his bachelor’s degree from Beihang University in 2021 and his master’s degree from Peking University in 2024. His research interests include diffusion models, visual tokenization, image copy detection, and optimization. He has published in top-tier conferences such as ICML and ICCV.
  \vspace{-1.5cm}
\end{IEEEbiography}
\vspace{-0.9cm}
\begin{IEEEbiography}
[{\includegraphics[width=1.0in,height=1.30in]{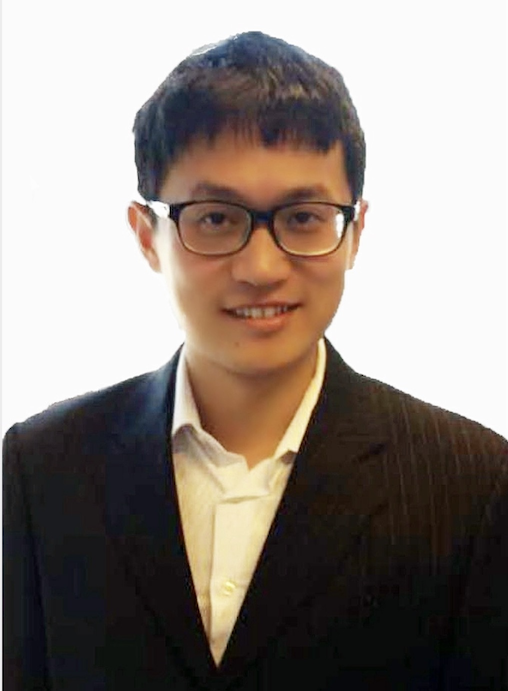}}]
{Dr. Yi Yang} (Senior Member, IEEE) is a distinguished Professor with the college of computer science and technology, Zhejiang University. He has authored over 200 papers in top-tier journals and conferences. His papers have received over 70,000 citations, with an H-index of 128. He has received more than 10 international awards in the field of AI, such as the Zhejiang Provincial Science Award First Prize, the Australian Research Council Discovery Early Career Research Award, the Australian Computer Society Gold Digital Disruptor Award, and the Google Faculty Research Award.
  \vspace{-1.5cm}
\end{IEEEbiography}

\clearpage
\setcounter{section}{0}
\setcounter{table}{0}
\setcounter{figure}{0}
\renewcommand{\thesection}{S\arabic{section}}
\renewcommand{\thetable}{S\arabic{table}}
\renewcommand{\thefigure}{S\arabic{figure}}

\begin{center}
{\Large\textbf{Supplementary Material}}
\end{center}
\vspace{-4mm}

\begin{figure*}[p]
\hspace*{-0.7cm}
\begin{forest}
  for tree={
  forked edges,
  grow=east,
  reversed=true,
  anchor=base west,
  parent anchor=east,
  child anchor=west,
  base=middle,
  font=\small,
  rectangle,
  draw=none, 
  edge={black!100, line width=0.4pt},
  rounded corners=2pt,
  minimum width=14em,
  s sep=4pt,
  inner xsep=3pt,
  inner ysep=2pt,
  scale=0.759,
  },
  [Replication,leaf5, fill=gray!10,text width=12em, fill opacity=1,draw=none, minimum width=10em
    [Unveiling ,leaf4,edge={black!100, line width=0.4pt}, fill=myblue, minimum height=1.2em ,draw=none,
        [Prompting,leaf3,edge={black!100, line width=0.4pt},fill=myblue,  draw=none
            [Specific,leaf2,fill=myblue,  draw=none,name=sn0
                 [  \cite{somepalli2023diffusion,carlini2023extracting,webster2023reproducible,naseh2023understanding,wang2023stronger,qu2023unsafe,brack2023distilling,wu2023proactive,leotta2023not,naik2023social}
                 , leaf1,no edge,fill=myblue ,draw=none,name=sn1
                 ]
            ]
            [Implicit,leaf2,fill=myblue, draw=none,name=sn2
                 [  \cite{zhang2024copyright,wen2024hard,naseh2023understanding}
                 , leaf1,no edge,fill=myblue ,draw=none,name=sn3
                 ]
            ]
        ]
        [Membership inference ,leaf3,edge={black!100, line width=0.4pt},fill=myblue, draw=none
            [White-box,leaf2, fill=myblue, draw=none,name=sn4
                 [  \cite{hu2023membership,matsumoto2023membership,pang2023white,duan2024membership,kong2023efficient}
                 , leaf1,no edge,fill=myblue ,draw=none,name=sn5
                 ]
            ]
            [Black-box ,leaf2,fill=myblue, draw=none,name=sn6
                 [  \cite{wu2022membership,matsumoto2023membership,fu2024probabilistic,laszkiewicz2023set,zhang2024generated,pang2023black,dubinski2024towards}
                 , leaf1,no edge,fill=myblue ,draw=none,name=sn7
                 ]
            ]
        ]
        [Similarity retrieval, leaf3,edge={black!100, line width=0.4pt}, fill=myblue,  draw=none,
            [Content similarity, leaf2,fill=myblue, draw=none,name=sn8
                 [  \cite{somepalli2023diffusion,bralios2024generation,rahman2024frame,zhou2023copyscope,aboutalebi2023deepfakeart,wu2024cgi}
                 , leaf1,no edge,fill=myblue ,draw=none,name=sn9
                 ]
            ]
            [Style similarity, leaf2,fill=myblue  , draw=none,name=sn10
                 [  \cite{casper2023measuring,somepalli2024measuring,wang2024AnyPattern}
                 , leaf1,no edge,fill=myblue ,draw=none,name=sn11
                 ]
            ]
        ]
        [Watermarking, leaf3, fill=myblue, draw=none, name=sl2
      [,no edge,[{\cite{wang2023diagnosis,cui2023diffusionshield,cui2023ft,luo2023steal}}, leaf1, no edge,fill=myblue, draw=none, name=sl3]]
    ]
        [Proactive replication, leaf3,edge={black!100, line width=0.4pt}, fill=myblue,  draw=none
            [Fine-tuning, leaf2,fill=myblue, draw=none,name=sn12
                 [  \cite{ruiz2023dreambooth,gal2022image, alaluf2023neural,arar2023domain,shah2023ziplora,chen2024subject,zhang2023inversion,jones2024customizing,kumari2023multi,gal2023encoder}
                 , leaf1,no edge,fill=myblue ,draw=none,name=sn13
                 ]
            ]
            [Training-free, leaf2,fill=myblue  , draw=none,name=sn14
                 [  \cite{ma2023subject,shi2023instantbooth}
                 , leaf1,no edge,fill=myblue ,draw=none,name=sn15
                 ]
            ]
        ]
        [Novel perspectives, leaf3,edge={black!100, line width=0.4pt}, fill=myblue,  draw=none
            [Magnitude of noise, leaf2,fill=myblue, draw=none,name=sn16
                 [  \cite{wen2024detecting}
                 , leaf1,no edge,fill=myblue ,draw=none,name=sn17
                 ]
            ]
            [Training data attribution, leaf2,fill=myblue  , draw=none,name=sn18
                 [  \cite{wang2023evaluating,georgiev2023journey}
                 , leaf1,no edge,fill=myblue ,draw=none,name=sn19
                 ]
            ]
            [Cross attention, leaf2,fill=myblue  , draw=none,name=sn20
                 [  \cite{ren2024unveiling}
                 , leaf1,no edge,fill=myblue ,draw=none,name=sn21
                 ]
            ]
            [Fine-tune to leak, leaf2,fill=myblue  , draw=none,name=sn22
                 [  \cite{li2024shake}
                 , leaf1,no edge,fill=myblue ,draw=none,name=sn23
                 ]
            ]
            [Overfitted MAE, leaf2,fill=myblue  , draw=none,name=sn24
                 [  \cite{taghanaki2024detecting}
                 , leaf1,no edge,fill=myblue ,draw=none,name=sn25
                 ]
            ]
            [Property inference, leaf2,fill=myblue  , draw=none,name=sn26
                 [  \cite{wang2024property}
                 , leaf1,no edge,fill=myblue ,draw=none,name=sn27
                 ]
            ]
        ]
    ]
    [Understanding, leaf4,edge={black!100, line width=0.4pt}, fill=mygreen, minimum height=1.2em, draw=none
        [Data ,leaf3,edge={black!100, line width=0.4pt}, fill=mygreen, minimum height=1.2em, draw=none
            [Insufficient training data,leaf2,fill=mygreen, draw=none,name=sm0
                 [  \cite{gu2023memorization}
                 , leaf1,no edge,fill=mygreen ,draw=none,name=sm1
                 ]
            ]
            [Image duplication,leaf2,fill=mygreen, draw=none,name=sm2
                 [  \cite{ chen2024towards,somepalli2023diffusion,somepalli2023understanding}
                 , leaf1,no edge,fill=mygreen ,draw=none,name=sm3
                 ]
            ]
            [Misleading captions, leaf2,fill=mygreen  , draw=none,name=sm4
                 [  \cite{somepalli2023diffusion,naseh2023memory,chen2024towards,somepalli2023understanding,gu2023memorization}
                 , leaf1,no edge,fill=mygreen ,draw=none,name=sm5
                 ]
            ]
            [Data types, leaf2,fill=mygreen  , draw=none,name=sm6
                 [  \cite{janolkar2023outliers}
                 , leaf1,no edge,fill=mygreen ,draw=none,name=sm7
                 ]
            ]
        ]
        [Methods, leaf3,edge={black!100, line width=0.4pt}, fill=mygreen, minimum height=1.2em, draw=none
            [Deterministic sampler,leaf2,fill=mygreen, draw=none,name=sm8
                 [  \cite{ yi2023generalization,zhang2023emergence}
                 , leaf1,no edge,fill=mygreen ,draw=none,name=sm9
                 ]
            ]
            [Model capacity,leaf2,fill=mygreen, draw=none,name=sm10
                 [  \cite{ somepalli2023understanding,chen2024towards}
                 , leaf1,no edge,fill=mygreen ,draw=none,name=sm11
                 ]
            ]
            [New metrics,leaf2,fill=mygreen, draw=none,name=sm12
                 [  \cite{ jiralerspong2023feature,jagielski2022measuring,li2024good}
                 , leaf1,no edge,fill=mygreen ,draw=none,name=sm13
                 ]]
        ]
        [Theory,leaf3,edge={black!100, line width=0.4pt}, fill=mygreen, minimum height=1.2em, draw=none
            [Near access-freeness ,leaf2,fill=mygreen, draw=none,name=sm14
                 [  \cite{vyas2023provable}
                 , leaf1,no edge,fill=mygreen ,draw=none,name=sm15
                 ]]
            [Dichotomy ,leaf2,fill=mygreen, draw=none,name=sm16
                 [  \cite{yoon2023diffusion}
                 , leaf1,no edge,fill=mygreen ,draw=none,name=sm17
                 ]]
            [Geometry-adaptive,leaf2,fill=mygreen, draw=none,name=sm18
                 [  \cite{kadkhodaie2023generalization}
                 , leaf1,no edge,fill=mygreen ,draw=none,name=sm19
                 ]]
            [Data-(in)dependent,leaf2,fill=mygreen, draw=none,name=sm20
                 [  \cite{li2023generalization}
                 , leaf1,no edge,fill=mygreen ,draw=none,name=sm21
                 ]]
            [Mutual information,leaf2,fill=mygreen, draw=none,name=sm22
                 [  \cite{yi2023generalization}
                 , leaf1,no edge,fill=mygreen ,draw=none,name=sm23
                 ]]
            [Creativity,leaf2,fill=mygreen, draw=none,name=sm24
                 [  \cite{wang2024can}
                 , leaf1,no edge,fill=mygreen ,draw=none,name=sm25
                 ]]
        ]
    ]
    [Mitigation,leaf4,edge={black!100, line width=0.4pt}, fill=mypurple, minimum height=1.2em, draw=none
        [Training data optimization,leaf3,edge={black!100, line width=0.4pt}, fill=mypurple, minimum height=1.2em, draw=none
            [Deduplication ,leaf2,fill=mypurple, draw=none,name=sj0
                 [  \cite{webster2023duplication,abbas2023semdedup,liao2022dataset,chen2024towards,somepalli2023understanding,li2024mitigate}
                 , leaf1,no edge,fill=mypurple ,draw=none,name=sj1
                 ]]
            [Protection ,leaf2,fill=mypurple, draw=none,name=sj2
                 [  \cite{ zheng2023understanding,zhao2023unlearnable,xue2023toward,zhu2024watermark,liang2023mist,van2023anti,liu2024toward,wang2023simac,xu2024perturbing, ye2023duaw,ma2023generative,tan2023pag,shan2023glaze,rhodes2023my,liang2023adversarial,dockhorn2023differentially,lyu2023differentially,amiridifferential,luo2024mpcpa,lebensold2024dp,qin2023destruction,zhao2023can,li2024va3,cao2023impress}
                 , leaf1,no edge,fill=mypurple ,draw=none,name=sj3
                 ]]
            [Purification ,leaf2,fill=mypurple, draw=none,name=sj4
                 [  \cite{gokaslan2023commoncanvas,abrahamsen2023inventing}
                 , leaf1,no edge,fill=mypurple ,draw=none,name=sj5
                 ]]
            [Corruption ,leaf2,fill=mypurple, draw=none,name=sj6
                 [  \cite{daras2023ambient,daras2024consistent}
                 , leaf1,no edge,fill=mypurple ,draw=none,name=sj7
                 ]]
        ]
        [Machine unlearning,leaf3,edge={black!100, line width=0.4pt}, fill=mypurple, minimum height=1.2em,draw=none,name=ss1
                 [,no edge,[  \cite{kumari2023ablating,zhang2023forget,wu2024erasediff,ren2024unveiling,hong2024all,bui2024removing,gandikota2023erasing,liu2024implicit,huang2023receler,fan2023salun,heng2023selective,li2024machine,das2024espresso,pham2024robust,yang2024pruning,li2024safegen,zhou2024plug,gandikota2024unified,kim2023towards,zhao2024separable,lu2024mace,xiong2024editing,zhang2024unlearncanvas,pham2023circumventing,zhang2023generate,tsai2023ring,petsiuk2024concept}
                 , leaf1,no edge,fill=mypurple ,draw=none,name=ss2
                 ]]]
        [Prompt disturbing,leaf3,edge={black!100, line width=0.4pt}, fill=mypurple, minimum height=1.2em, draw=none,name=st1
                 [,no edge,[  \cite{wen2024detecting,schramowski2023safe,dong2023towards,ni2023degeneration,li2023get}
                 , leaf1,no edge,fill=mypurple ,draw=none,name=st2
                 ]]]
        [Novel perspectives,leaf3,edge={black!100, line width=0.4pt}, fill=mypurple, minimum height=1.2em, draw=none
            [Composition ,leaf2,fill=mypurple, draw=none,name=sj8
                 [  \cite{golatkar2023training}
                 , leaf1,no edge,fill=mypurple ,draw=none,name=sj9
                 ]]
            [Model immunizing ,leaf2,fill=mypurple, draw=none,name=sj10
                 [  \cite{zheng2023imma}
                 , leaf1,no edge,fill=mypurple ,draw=none,name=sj11
                 ]]
            [Low-rank adaptation ,leaf2,fill=mypurple, draw=none,name=sj12
                 [  \cite{luo2024privacy}
                 , leaf1,no edge,fill=mypurple ,draw=none,name=sj13
                 ]]
            [Despecification guidance ,leaf2,fill=mypurple, draw=none,name=sj14
                 [  \cite{chen2024towards}
                 , leaf1,no edge,fill=mypurple ,draw=none,name=sj15
                 ]]
        ]
    ]
    [Influence,leaf4,edge={black!100, line width=0.4pt}, fill=myyellow, draw=none,name=sp0
                [Regulation ,leaf2,edge={black!100, line width=0.4pt},fill=myyellow, draw=none,name=sm26
                 [,no edge, [  \cite{zirpoli2023generative,lee2023ai,henderson2023foundation,samuelson2023generative,lee2024talkin,murray2023generative,sag2023copyright,lemley2023generative,wang2023analyzing,cooper2024files,peukert2024economics,elkin2023can}
                 , leaf1,no edge,fill=myyellow ,draw=none,name=sm27
                 ]]]
            [Art ,leaf2,fill=myyellow, draw=none,name=sm28
                 [,no edge, [  \cite{rudolf2024understanding,jiang2023ai,aiart,ghosh2022can,gabrys2023ai,moayeri2024rethinking,crawford2022legal}
                 , leaf1,no edge,fill=myyellow ,draw=none,name=sm29
                 ]]]
            [Society,leaf2,fill=myyellow, draw=none,name=sm30
                 [,no edge, [  \cite{ luccioni2023stable,perera2023analyzing,zhang2023auditing,wu2023stable}
                 , leaf1,no edge,fill=myyellow ,draw=none,name=sm31
                 ]]]
             [Healthcare,leaf2,fill=myyellow, draw=none,name=sp1
                 [,no edge, [  \cite{ dar2023investigating,dar2024effect,akbar2023beware,dar2024unconditional,usman2024brain,fernandez2023privacy}
                 , leaf1,no edge,fill=myyellow ,draw=none,name=sp2
                 ]]]
        ]]
  ]
  \draw[-, black, line width=0.4pt] (sn0) -- (sn1);
  \draw[-, black, line width=0.4pt] (sn2) -- (sn3);
  \draw[-, black, line width=0.4pt] (sn4) -- (sn5);
  \draw[-, black, line width=0.4pt] (sn6) -- (sn7);
  \draw[-, black, line width=0.4pt] (sn8) -- (sn9);
  \draw[-, black, line width=0.4pt] (sn10) -- (sn11);
  \draw[-, black, line width=0.4pt] (sn12) -- (sn13);
  \draw[-, black, line width=0.4pt] (sn14) -- (sn15);
  \draw[-, black, line width=0.4pt] (sn16) -- (sn17);
  \draw[-, black, line width=0.4pt] (sn18) -- (sn19);
  \draw[-, black, line width=0.4pt] (sn20) -- (sn21);
  \draw[-, black, line width=0.4pt] (sn22) -- (sn23);
  \draw[-, black, line width=0.4pt] (sn24) -- (sn25);
  \draw[-, black, line width=0.4pt] (sn26) -- (sn27);
  \draw[-, black, line width=0.4pt] (sm0) -- (sm1);
  \draw[-, black, line width=0.4pt] (sm2) -- (sm3);
  \draw[-, black, line width=0.4pt] (sm4) -- (sm5);
  \draw[-, black, line width=0.4pt] (sm6) -- (sm7);
  \draw[-, black, line width=0.4pt] (sm8) -- (sm9);
  \draw[-, black, line width=0.4pt] (sm10) -- (sm11);
  \draw[-, black, line width=0.4pt] (sm12) -- (sm13);
  \draw[-, black, line width=0.4pt] (sm14) -- (sm15);
  \draw[-, black, line width=0.4pt] (sm16) -- (sm17);
  \draw[-, black, line width=0.4pt] (sm18) -- (sm19);
  \draw[-, black, line width=0.4pt] (sm20) -- (sm21);
  \draw[-, black, line width=0.4pt] (sm22) -- (sm23);
  \draw[-, black, line width=0.4pt] (sm24) -- (sm25);
  \draw[-, black, line width=0.4pt] (sm26) -- (sm27);
  \draw[-, black, line width=0.4pt] (sm28) -- (sm29);
  \draw[-, black, line width=0.4pt] (sm30) -- (sm31);
  \draw[-, black, line width=0.4pt] (sj0) -- (sj1);
  \draw[-, black, line width=0.4pt] (sj2) -- (sj3);
  \draw[-, black, line width=0.4pt] (sj4) -- (sj5);
  \draw[-, black, line width=0.4pt] (sj6) -- (sj7);
  \draw[-, black, line width=0.4pt] (sj8) -- (sj9);
  \draw[-, black, line width=0.4pt] (sj10) -- (sj11);
  \draw[-, black, line width=0.4pt] (sj12) -- (sj13);
  \draw[-, black, line width=0.4pt] (sj14) -- (sj15);
  \draw[-, black, line width=0.4pt] (sl2) -- (sl3);
  \draw[-, black, line width=0.4pt] (sp1) -- (sp2);
  \draw[-, black, line width=0.4pt] (st1) -- (st2);
  \draw[-, black, line width=0.4pt] (ss1) -- (ss2);
\end{forest}
\caption{Categorization of the literature on replication in visual diffusion models: unveiling, understanding, mitigation, and its influence.}
\label{Fig: overview}
\end{figure*}

\begin{addedblock}
\section{Replication in Video and 3D Diffusion Models}
\label{sec:video3d}
While the majority of existing work targets image diffusion models, the replication phenomenon is equally concerning in emerging video and 3D diffusion models. We briefly outline the unique challenges and early findings in these domains.

\noindent\textbf{Video diffusion models.}
Video diffusion models \cite{ho2022video,singer2023make,blattmann2023svd,openai2024sora,polyak2024moviegen,yang2024cogvideox} extend image generation to the temporal domain and have attracted enormous research interest in 2024--2025. The training data for these models typically consists of video clips sourced from the web, where temporal redundancy is far greater than spatial redundancy in images: the same scene or person often appears across thousands of frames or even multiple clips. This redundancy amplifies replication risk along two dimensions. First, \textit{clip-level replication}\added{, \ie, the reproduction of an entire short video sequence rather than a single frame,} is harder to detect than single-frame replication because standard copy-detection pipelines operate frame-independently. \cite{rahman2024frame} takes an early step in this direction by adapting frame-level similarity retrieval to the video domain, and \cite{videomem2025} confirms that memorization is widespread across all tested video diffusion models, but a principled clip-level replication metric remains an open problem. Second, \textit{subject identity replication}\added{, \ie, the reproduction of a specific person's face or body across generated frames,} poses heightened privacy risks, since videos carry richer biometric information than still images.

\noindent\textbf{3D diffusion models.}
3D diffusion models \cite{poole2023dreamfusion,shi2024mvdream} trained on geometric datasets (\textit{e.g.}, ShapeNet \cite{chang2015shapenet}) face a distinct form of replication: not only can surface textures be replicated (analogous to content-level image replication), but the underlying geometry, such as mesh topology, volume structure, can also be memorized and reproduced. This compounds the risk: a generated 3D asset may replicate copyrighted geometry even when its texture appears novel. Existing replication metrics defined for 2D images (\textit{e.g.}, SSCD \cite{pizzi2022self}) do not directly apply to 3D representations, necessitating new distance metrics for point clouds, meshes, and implicit neural fields. Recent work \cite{memorization3d2025} provides the first empirical study of memorization in 3D shape generation, finding that memorization depends on data modality, increases with data diversity and finer-grained conditioning, and can be mitigated by rotation augmentation.
\end{addedblock}

\begin{addedblock}
\section{Detailed Benchmark Tables}

\subsection{Membership Inference Attacks Benchmark}

\begin{table*}[h]
\caption{Comprehensive MIA benchmark across diffusion models. ``BB'' = black-box; ``WB'' = white-box. \textbf{Upper block}: controlled settings (small datasets, models trained from scratch). \textbf{Lower block}: realistic settings (pre-trained Stable Diffusion on LAION). The AUC drop from controlled to realistic settings reveals a evaluation gap in the MIA literature.}
\label{tab:mia}
\setlength{\tabcolsep}{3.8pt}
\renewcommand{\arraystretch}{1}
\begin{tabular*}{\textwidth}{@{\extracolsep{\fill}}llllccl}
\hline
\textbf{Method} & \textbf{Type} & \textbf{Model / Setting} & \textbf{Dataset} & \textbf{AUC$\uparrow$} & \textbf{TPR@1\%$\uparrow$} & \textbf{Reference} \\
\hline
\multicolumn{7}{l}{\textit{Controlled settings (trained from scratch or fine-tuned on small data)}} \\
\hline
SecMI & BB & DDPM & CIFAR-10 & 0.881 & 9.1\% & Duan~\textit{et al.}~\cite{duan2023diffusion} \\
PIA & BB & DDPM & CIFAR-10 & 0.885 & 13.7\% & Kong~\textit{et al.}~\cite{kong2023efficient} \\
PIAN & BB & DDPM & CIFAR-10 & 0.878 & 31.2\% & Kong~\textit{et al.}~\cite{kong2023efficient} \\
PFAMI & BB & DDPM & CelebA-64 & 0.986 & 50.2\% & Fu~\textit{et al.}~\cite{fu2024probabilistic} \\
GSA & WB & DDPM & CIFAR-10 & 1.000 & 100\% & Pang~\textit{et al.}~\cite{pang2023white} \\
Noise-as-Probe & BB & Fine-tuned SD & MS-COCO & 0.905 & 21.8\% & Choi~\textit{et al.}~\cite{choi2025noiseasprobe} \\
Li~\textit{et al.} & BB & Fine-tuned SD & CelebA-Dialog & 0.930 & 60.0\% & Li~\textit{et al.}~\cite{li2025blackboxndss} \\
\hline
\multicolumn{7}{l}{\textit{Realistic settings (pre-trained on LAION)}} \\
\hline
SecMI & BB & SD v1.5 & LAION & 0.523 & 1.3\% & Tang~\textit{et al.}~\cite{duan2024membership} \\
PIA & BB & SD v1.5 & LAION & 0.535 & 1.3\% & Tang~\textit{et al.}~\cite{duan2024membership} \\
PFAMI & BB & SD v1.5 & LAION & 0.510 & 1.6\% & Tang~\textit{et al.}~\cite{duan2024membership} \\
Dubinski~\textit{et al.} & BB & SD v1.4 & LAION-mi & 0.521 & 2.5\% & Dubinski~\textit{et al.}~\cite{dubinski2024towards} \\
GSA & WB & SD v1.5 & LAION & 1.000 & 100\% & Tang~\textit{et al.}~\cite{duan2024membership} \\
\hline
\textit{Ours} (loss-based) & WB & SD v1.5 & Controlled & 0.612 & 14.0\% & --- \\
\textit{Ours} (gradient-norm) & WB & SD v1.5 & Controlled & 0.617 & 32.0\% & --- \\
\hline
\end{tabular*}
\end{table*}

\subsection{Memorization Extraction and Detection}

\begin{table*}[h]
\caption{Compilation of published memorization extraction and detection results across diffusion models. ``Extraction'' measures actual training data recovery; ``Detection'' measures the ability to identify memorized prompts/images.}
\label{tab:memorization_rates}
\setlength{\tabcolsep}{3.8pt}
\renewcommand{\arraystretch}{1}
\begin{tabular*}{\textwidth}{@{\extracolsep{\fill}}llllcl}
\hline
\textbf{Study} & \textbf{Model} & \textbf{Task} & \textbf{Metric} & \textbf{Result} & \textbf{Key finding} \\
\hline
Carlini~\textit{et al.}~\cite{carlini2023extracting} & SD v1.4 & Extraction & L$_2$ $<$ 0.15 & 109 / 175M & Low rate but non-zero verbatim copies \\
Carlini~\textit{et al.}~\cite{carlini2023extracting} & Imagen & Extraction & L$_2$ $<$ 0.15 & 2.3\% (23/1K prompts) & Closed-source models also memorize \\
Carlini~\textit{et al.}~\cite{carlini2023extracting} & CIFAR-10 DDPM & Extraction & L$_2$ match & 2.5\% of dataset & Unconditional models memorize less \\
Somepalli~\textit{et al.}~\cite{somepalli2023diffusion} & SD (LAION-Aes) & Retrieval & SSCD $\geq$ 0.5 & 1.88\% of generations & Object-level copies in web-trained models \\
Somepalli~\textit{et al.}~\cite{somepalli2023understanding} & SD 2.1 & Retrieval & SSCD $\geq$ 0.5 & 1.2\% of generations & Text conditioning amplifies replication \\
Webster~\cite{webster2023reproducible} & SD v1 & Extraction & MSE $<$ 0.12 & 71 / 500 prompts & 2000$\times$ more efficient than~\cite{carlini2023extracting} \\
Webster~\cite{webster2023reproducible} & SD v2 & Extraction & MSE $<$ 0.12 & 4 / 500 prompts & Deduplication reduces verbatim copies \\
Gu~\textit{et al.}~\cite{gu2023memorization} & CIFAR-10 DDPM & EMM & Memorization ratio & $>$60\% at $|D|$=16K & EMM grows with resolution and capacity \\
Wen~\textit{et al.}~\cite{wen2024detecting} & SD & Detection & AUC & 0.999 (32 gen, 10 steps) & Near-perfect memorization detection \\
Wen~\textit{et al.}~\cite{wen2024detecting} & SD & Detection & AUC & 0.960 (1 gen, 1 step) & Single-query detection still effective \\
\hline
\textit{Ours} & SD v1.5 & Self-sim & CLIP cos (same prompt) & 0.906 & Prompt-driven replication dominates \\
\textit{Ours} & SD v1.5 & Self-sim & CLIP cos (diff prompt) & 0.574 & Cross-prompt similarity much lower \\
\hline
\end{tabular*}
\end{table*}

\subsection{Data Protection and Differential Privacy}

\begin{table*}[h]
\caption{Data protection and differential privacy methods for diffusion models. \textbf{Upper}: adversarial perturbation methods; protection success measures the fraction of images for which mimicry is effectively prevented. \textbf{Lower}: differential privacy methods; FID measures generation quality at a given privacy budget $\varepsilon$.}
\label{tab:protection}
\setlength{\tabcolsep}{3.8pt}
\renewcommand{\arraystretch}{1}
\begin{tabular*}{\textwidth}{@{\extracolsep{\fill}}llccl}
\hline
\multicolumn{5}{l}{\textit{Adversarial protection methods}} \\
\hline
\textbf{Method} & \textbf{Target} & \textbf{Metric} & \textbf{Result} & \textbf{Reference} \\
\hline
Glaze & Style & Success rate & $>$92\% & Shan~\textit{et al.}~\cite{shan2023glaze} \\
Glaze & Style & Artist-rated & 93.3\% & Shan~\textit{et al.}~\cite{shan2023glaze} \\
LightShed & Anti-Glaze & Detection acc. & 99.98\% & Foerster~\textit{et al.}~\cite{foerster2025lightshed} \\
Anti-DB (ASPL) & Identity & FDFR$\uparrow$ & 0.63 & Van Le~\textit{et al.}~\cite{van2023anti} \\
Anti-DB (ASPL) & Identity & ISM$\downarrow$ & 0.33 & Van Le~\textit{et al.}~\cite{van2023anti} \\
Mist & Content & FID (disruption) & 91.0 & Liang~\textit{et al.}~\cite{liang2023mist} \\
AdvDM & Content & FID (disruption) & 98.3 & Liang~\textit{et al.}~\cite{liang2023adversarial} \\
\hline
\multicolumn{5}{l}{\textit{Differential privacy methods (lower FID = better quality)}} \\
\hline
\textbf{Method} & \textbf{Dataset} & $\boldsymbol{\varepsilon}$ & \textbf{FID$\downarrow$} & \textbf{Reference} \\
\hline
DPDM & MNIST & 10 & 5.01 & Dockhorn~\textit{et al.}~\cite{dockhorn2023differentially} \\
DP-LDM & CIFAR-10 & 10 & 8.4 & Lebensold~\textit{et al.}~\cite{lebensold2024dp} \\
DP-LDM & CIFAR-10 & 1 & 22.9 & Lebensold~\textit{et al.}~\cite{lebensold2024dp} \\
DP-LoRA & CelebA-64 & 10 & 8.4 & Tsai~\textit{et al.}~\cite{tsai2025dplora} \\
DP-LoRA & CelebA-64 & 1 & 12.0 & Tsai~\textit{et al.}~\cite{tsai2025dplora} \\
\hline
\end{tabular*}
\vspace{-3mm}
\end{table*}

\subsection{Machine Unlearning Benchmark}

\begin{table*}[h]
\caption{Machine unlearning benchmark for diffusion models. \textbf{Upper}: UnlearnCanvas~\cite{zhang2024unlearncanvas} results on style and object unlearning (Stable Diffusion v1.4). UA = Unlearning Accuracy ($\uparrow$), IRA = In-domain Retainability ($\uparrow$), CRA = Cross-domain Retainability ($\uparrow$). \textbf{Lower}: generation quality and efficiency. No single method dominates across all metrics.}
\label{tab:unlearning}
\setlength{\tabcolsep}{3.8pt}
\renewcommand{\arraystretch}{1}
\begin{tabular*}{\textwidth}{@{\extracolsep{\fill}}lcccccccccc}
\hline
& \multicolumn{3}{c}{\textbf{Style Unlearning}} & \multicolumn{3}{c}{\textbf{Object Unlearning}} & & & \\
\cline{2-7}
\textbf{Method} & \textbf{UA$\uparrow$} & \textbf{IRA$\uparrow$} & \textbf{CRA$\uparrow$} & \textbf{UA$\uparrow$} & \textbf{IRA$\uparrow$} & \textbf{CRA$\uparrow$} & \textbf{FID$\downarrow$} & \textbf{Time (s)} & \textbf{Reference} \\
\hline
ESD & 98.6 & 81.0 & 94.0 & 92.2 & 55.8 & 44.2 & 65.6 & 6163 & Gandikota~\textit{et al.}~\cite{gandikota2023erasing} \\
FMN & 88.5 & 56.8 & 46.6 & 45.6 & 90.6 & 73.5 & 131.4 & 350 & Zhang~\textit{et al.}~\cite{zhang2023forget} \\
UCE & 98.4 & 60.2 & 47.7 & 94.3 & 39.4 & 34.7 & 182.0 & 434 & Gandikota~\textit{et al.}~\cite{gandikota2024unified} \\
CA & 60.8 & 96.0 & 92.7 & 46.7 & 90.1 & 82.0 & 54.2 & 734 & Kumari~\textit{et al.}~\cite{kumari2023ablating} \\
SalUn & 86.3 & 90.4 & 95.1 & 86.9 & 96.4 & 99.6 & 61.1 & 667 & Fan~\textit{et al.}~\cite{fan2023salun} \\
\hline
\end{tabular*}
\end{table*}
\end{addedblock}

\begin{addedblock}
\vspace{3mm}
\section{Cross-Cutting Conclusions on Data Properties}
\label{sec:supp_data}
From the above findings, we draw several cross-cutting conclusions about how dataset properties influence the degree of replication.
\begin{enumerate}[leftmargin=*]
    \item \textit{Scale matters, but non-linearly.} \cite{gu2023memorization} shows that the replication ratio decreases as dataset size grows, but the relationship is sublinear: halving the dataset more than doubles the per-image replication probability. In the extreme small-data regime (as in medical imaging \cite{dar2023investigating,usman2024brain}), replication becomes near-certain.
    \item \textit{Duplication rate is a stronger predictor than raw size.} \cite{somepalli2023understanding} demonstrates that a large dataset with high duplication (common in web-scraped corpora) can exhibit higher replication rates than a smaller but carefully deduplicated dataset. This suggests that data cleaning, not merely data collection, is the primary lever for controlling replication.
    \item \textit{Caption specificity interacts with duplication.} \cite{naseh2023memory,gu2023memorization} find that unique or inaccurate captions exacerbate replication when paired with duplicated images, because the model cannot distribute its capacity across diverse captions. Conversely, diverse captions partially compensate for moderate image duplication.
    \item \textit{Data type determines which semantic level is replicated.} Web images with repeated subjects (\textit{e.g.}, celebrity photos) predominantly cause content-level replication; stylistically homogeneous artistic datasets (\textit{e.g.}, a single artist's portfolio) cause style-level replication; and datasets with strong demographic skew cause concept-level replication via biased associations.
\end{enumerate}
These insights collectively motivate the mitigation strategies reviewed in Section~\ref{sec:mitigate}: deduplication targets factor (ii), caption diversification targets factor (iii), and differential privacy targets factor (i) by bounding the per-sample influence.
\end{addedblock}

\begin{addedblock}
\section{Limitations and Deployment Challenges of Machine Unlearning}
\label{sec:supp_unlearn_limits}
\noindent\textbf{Limitations and deployment challenges.}
Although the machine unlearning methods reviewed above show promise, several fundamental challenges remain before they can be reliably deployed in practice.
\begin{enumerate}[leftmargin=*]
    \item \textit{Catastrophic forgetting.} Aggressive unlearning of a target concept often degrades the model's ability to generate semantically related but benign content. For example, erasing a specific artist's style may impair generation of broader artistic styles. UnlearnCanvas \cite{zhang2024unlearncanvas} empirically demonstrates that many methods fail to balance retainability with erasure.
    \item \textit{Incomplete unlearning.} Memorized content may persist in model weights even after targeted unlearning: \cite{pham2023circumventing} shows that learned word embeddings can re-elicit erased concepts without modifying the model's weights, and \cite{petsiuk2024concept} demonstrates that combining multiple prompts can reconstruct erased concept vectors. This implies that current unlearning methods offer only a soft guarantee.
    \item \textit{Scalability constraints.} Most methods require access to the original training data or per-sample gradients. In API-only deployment settings, neither is available. Furthermore, MACE \cite{lu2024mace} and EMCID \cite{xiong2024editing} show that even scaling to 1,000 concepts introduces measurable degradation in unrelated generation quality.
    \item \textit{Verification difficulty.} There is no reliable protocol to certify that unlearning is complete. Evaluation typically relies on held-out prompts, which can be circumvented by adversarially designed queries \cite{tsai2023ring,zhang2023generate}. A standardized auditing framework, analogous to security penetration testing, is needed but absent.
    \item \textit{Regulatory uncertainty.} It remains legally unclear whether successful machine unlearning satisfies ``right to be forgotten'' requirements under GDPR \cite{voigt2017eu} or analogous legislation, since no court has yet adjudicated whether computational forgetting is equivalent to data deletion.
\end{enumerate}

To quantify these trade-offs, Table~\ref{tab:unlearning} compiles published results from the UnlearnCanvas benchmark~\cite{zhang2024unlearncanvas} and the Holistic Unlearning Benchmark (HUB)~\cite{moon2025hub}, which evaluate representative methods across unlearning accuracy (UA), in-domain retainability (IRA), cross-domain retainability (CRA), and generation quality (FID). Key observations: (i)~ESD~\cite{gandikota2023erasing} and UCE~\cite{gandikota2024unified} achieve the highest UA ($>$98\%) but suffer significant IRA/CRA degradation; (ii)~SalUn~\cite{fan2023salun} offers the best retainability (CRA up to 99.6\%) at the cost of moderate UA; (iii)~CA~\cite{kumari2023ablating} preserves generation quality (FID 54.2) but has the weakest erasure effectiveness; (iv)~no single method dominates across all metrics, underscoring the need for task-specific selection.
\end{addedblock}

\section{Influence}\label{Medical}
After sequentially discussing the unveiling, understanding, and mitigation of replication in visual diffusion models, as outlined in Fig. \ref{Fig: overview}, this section focuses on its \textit{influence} in the real world.
Specifically, as illustrated in Fig. \ref{Fig: influence}, we focus on regulation, art, society, and healthcare.
This involves opinions from legal scholars, artists, sociologists, and doctors.

\subsection{Regulation}
\begin{figure*}[t]
    \centering
    \includegraphics[width=1\textwidth]{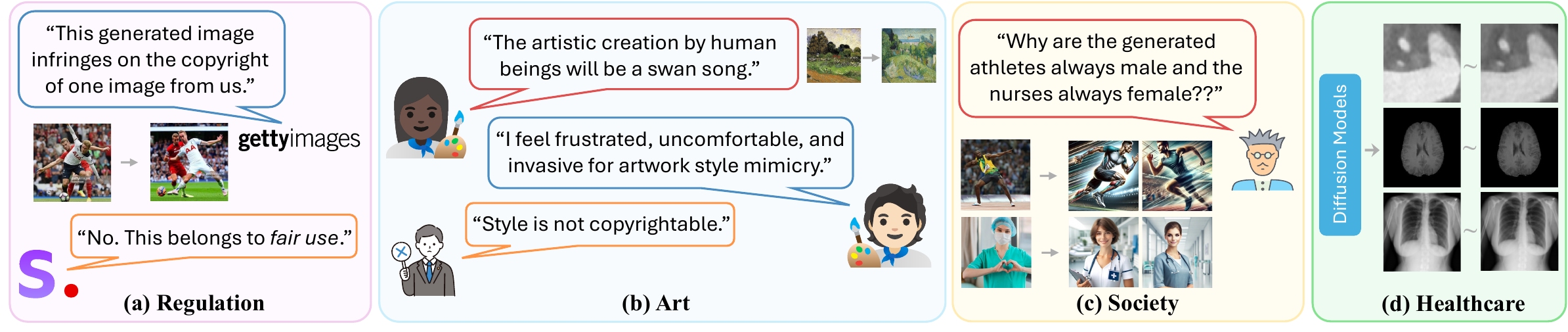}
    \vspace{-6mm}
    \caption{Illustrations of different influences of replication in visual diffusion models.}
    \label{Fig: influence}
\end{figure*}

As shown in Fig. \ref{Fig: influence} (a), training and generating processes in some visual diffusion models raise significant law issues due to the replication of copyrighted materials. As these models become more powerful and prevalent, an increasing number of legal scholars are focusing on this area. They primarily investigate how these models manage and utilize copyrighted materials during the creation process, along with the challenges and implications for the existing copyright law framework. For instance, they \cite{zirpoli2023generative,lee2023ai,henderson2023foundation,samuelson2023generative,lee2024talkin,lemley2023generative,murray2023generative,sag2023copyright} question whether using copyrighted works as training data for AI constitutes copyright infringement, whether AI-generated outputs are derivative works infringing on the original copyrights, and who owns the copyright for AI-generated works. Furthermore, \cite{lemley2023generative,wang2023analyzing} discuss the intricate infringement challenges that arise when generative AI models, particularly visual diffusion models, are trained using copyrighted materials without proper authorization. Additionally, \cite{cooper2024files} aims to define and clarify what constitutes replication from the perspective of copyright infringement; \cite{peukert2024economics} thoroughly explores the intersection of copyright law and economic principles in the context of rapid technological advancements; and the core idea of \cite{elkin2023can} is to evaluate whether privacy protection measures can align with and support copyright law.\par

\begin{table*}[t]
\caption{Comparative overview of international AI/copyright regulatory approaches relevant to diffusion model replication.}
\label{tab:regulation}
\setlength{\tabcolsep}{3.8pt}
\renewcommand{\arraystretch}{1}
\begin{tabular*}{\textwidth}{@{\extracolsep{\fill}}llll}
\hline
\textbf{Jurisdiction} & \textbf{Training-data exception} & \textbf{Opt-out recognized?} & \textbf{AI output copyrightable?} \\
\hline
\textbf{USA} & Fair use (case-by-case) & No statutory basis & No \cite{uspto2023ai} \\
\textbf{EU} & TDM exception (CDSM Art.~4, opt-out allowed) & Yes & Not as-of-right \\
\textbf{UK} & TDM exception (CDPA s.~29A; proposed expansion withdrawn) & Uncertain & Possible (CDPA s.~9(3)) \\
\textbf{China} & Broad exception; new Interim Measures (2023) require lawful data sourcing & Partially & Case-by-case \\
\textbf{Japan} & Broad TDM exception (Copyright Act Art.~30-4, non-enjoyment purpose) & Not required & Case-by-case \\
\hline
\end{tabular*}
\vspace{-3mm}
\end{table*}

\begin{addedblock}
\noindent\textbf{Unveiling through litigation.}
Several landmark lawsuits have acted as ``whistleblowers'' that expose how diffusion models replicate training data at scale. In \textit{Andersen v.\ Stability AI}~\cite{andersen2023stability}, a class of visual artists alleged that Stable Diffusion~\cite{rombach2022high} was trained on their copyrighted works without consent, and that its outputs constitute infringing derivative works. Similarly, \textit{Getty Images v.\ Stability AI}~\cite{getty2023stability} focuses on the near-verbatim replication of Getty's watermarked photographs, with watermark fragments visible in generated outputs serving as forensic evidence of content-level replication. These cases operationalize replication detection as a legal evidentiary tool. In November 2025, the UK High Court ruled in \textit{Getty Images v.\ Stability AI} \cite{gettyruling2025} that although the model is exposed to copyrighted works during training, it does not store the training data itself, rejecting the copyright infringement claim, a landmark decision with significant implications for the field. Meanwhile, the U.S. Copyright Office concluded \cite{uscopyrightoffice2025} that AI developers using copyrighted works to train models generating ``expressive content that competes with'' original works exceed the scope of fair use.\par

\noindent\textbf{Understanding applicable legal doctrines.}
The central doctrinal question varies by jurisdiction. \cite{sag2023copyright,lemley2023generative} analyze whether AI training constitutes fair use under US copyright law: while the transformative-use factor may favor developers, the commercial nature of outputs and market substitution weigh against it. \cite{henderson2023foundation} empirically demonstrates that foundation models can reproduce verbatim training data under adversarial prompting, strengthening the plaintiffs' case. In the EU, the Copyright in the Digital Single Market Directive~\cite{cdsm2019} introduced a text-and-data mining (TDM) exception (Article 4), but rights holders can opt out; the legal validity of mass opt-out mechanisms for AI training remains unsettled \cite{peukert2024economics}.\par

\noindent\textbf{Mitigation through regulatory compliance.}
Effective compliance strategies are emerging in response to this regulatory landscape:
\begin{enumerate}[leftmargin=*]
    \item \textit{Opt-out and licensing mechanisms.} Platforms such as DeviantArt (NoAI tag) and Adobe Stock (opt-in only) have begun implementing opt-out metadata standards. \cite{henderson2023foundation} proposes a structured approach to auditing whether such opt-outs are honored at training time.
    \item \textit{Training data disclosure.} The EU AI Act~\cite{euaiact2024} requires providers of general-purpose AI models to publish summaries of training data, enabling rights holders to identify unauthorized use \cite{guadamuz2025euaiact,quintais2025genai}. As of August 2025, every GPAI provider must maintain documentation of training data provenance and prove compliance with EU copyright rules. This creates an institutional incentive for replication auditing analogous to the methods surveyed in Section~\ref{Unveiling}.
    \item \textit{GDPR ``right to be forgotten.''} Personal data replicated by diffusion models falls under GDPR Article 17. Machine unlearning is the primary technical mechanism proposed to implement this right, though the legal sufficiency of computational forgetting remains untested in court.
\end{enumerate}

\noindent\textbf{International comparative analysis.}
The regulatory landscape is highly heterogeneous across jurisdictions, as summarized in Table~\ref{tab:regulation}. The key divergences concern (i) whether training on copyrighted data is prima facie infringing, (ii) whether opt-out mechanisms are legally recognized, and (iii) whether AI-generated outputs can themselves be copyrighted.\par

\end{addedblock}

Beyond the copyright issues, there are also privacy concerns and corresponding data protection regulations \cite{lyu2023pathway,cooper2023report}. The replication of data by visual diffusion models can pose significant privacy risks, especially when the models inadvertently replicate sensitive or personal data. This contravenes data protection regulations such as the General Data Protection Regulation (GDPR) \cite{voigt2017eu} in Europe, which mandates the protection of personal data with appropriate technical measures. Regulatory frameworks ensure that AI systems, particularly those trained on vast amounts of potentially sensitive data, comply with privacy regulations and do not retain or reproduce personal data
without consent.\par

The replication of biases in training data by AI models is another regulatory concern  \cite{lyu2023pathway,ghosh2023dual,bartlett2024generative}. Ensuring that diffusion models do not perpetuate or amplify biases present in the data they are trained on is crucial. Regulations enforce fairness, accountability, and transparency in AI systems to mitigate these issues. This could involve mandatory bias audits, transparency in data usage, and clear documentation of the data and methodologies used in training AI models.\par

\subsection{Art}
The influence of generative AI in art worlds presents both opportunities and challenges. These models have transformed the art market, personalizing the buying experience and enhancing the efficiency of curators in identifying trends and managing collections \cite{rudolf2024understanding}. Despite these advances, as shown in Fig. \ref{Fig: influence} (b), many artists fear that AI may threaten their jobs and dilute the authenticity of art by replicating styles and producing art without human involvement \cite{jiang2023ai,aiart}. This fuels the ongoing debate about whether art can exist without an artist \cite{ghosh2022can,gabrys2023ai}. Additionally, some researchers \cite{moayeri2024rethinking} are investigating artistic copyright infringements, underscoring the complex challenges in protecting intellectual property because artistic style itself is not copyrightable \cite{crawford2022legal}.

\begin{addedblock}
\noindent\textbf{Unveiling style replication in art.}
Several real-world controversies have emerged that concretely illustrate how diffusion models replicate artistic style.
(i) \textit{Hollie Mengert.} The illustrator Hollie Mengert became one of the first documented cases in which a diffusion model (NovelAI, which built on Stable Diffusion \cite{rombach2022high}) was used to generate images in a style closely matching her distinctive flat, graphic work. This case exemplifies style-level replication: no individual image was directly copied, but the statistical signature of her style, as measurable by style-similarity metrics \cite{somepalli2024measuring,wang2024AnyPattern}, was reproduced at will.
(ii) \textit{DeviantArt / NovelAI controversy.} In 2022, NovelAI released an image-generation model trained on ArtStation and DeviantArt images without artist consent. Artists organized mass protests by flooding ArtStation with ``No AI Art'' posts. This incident prompted ArtStation and DeviantArt to introduce opt-out mechanisms and ``AI-generated'' content labels.
(iii) \textit{Empirical style-replication measurement.} \cite{somepalli2024measuring} provides the first large-scale quantitative study of style replication, measuring cosine similarity in a style-aware feature space between LAION-trained model outputs and training images. They find that style replication is prevalent even when content is not directly copied, and that replication rate correlates with training dataset duplication frequency.\par

\noindent\textbf{Understanding artistic style vulnerability.}
Several factors make style-level replication particularly pervasive in artistic domains:
(i) \textit{Style is not copyrightable}, creating a legal gap \cite{crawford2022legal}: even if a model demonstrably replicates an artist's style, current copyright law in most jurisdictions provides no remedy.
(ii) \textit{Stylistically homogeneous training sets.} When a model is fine-tuned on a single artist's portfolio (\textit{e.g.}, via DreamBooth \cite{ruiz2023dreambooth}), the resulting model can generate near-perfect style replications on demand. \cite{moayeri2024rethinking} analyzes this phenomenon and questions whether style fine-tuning constitutes a form of creative appropriation that current legal frameworks fail to address.
(iii) \textit{CLIP-based training amplifies style coherence.} Because CLIP encodes stylistic associations in its embedding space, models trained with CLIP guidance tend to replicate style clusters present in the training distribution more faithfully than unconditional models.

\noindent\textbf{Mitigation through style protection.}
Mitigation strategies for artistic style replication mirror those for content replication but require style-aware tools:
(i) \textit{Style-adversarial perturbations (Glaze \cite{shan2023glaze}).} Glaze adds imperceptible perturbations to an artist's images before public release, shifting their style-space representation to confuse fine-tuning pipelines while remaining visually faithful to the artist's intent.
(ii) \textit{Opt-out metadata and content filtering.} Platforms now support opt-out flags (\textit{e.g.}, DeviantArt's NoAI tag), and some model providers filter style-specific fine-tuning requests.
(iii) \textit{Regulatory advocacy.} Artists have lobbied for amendments to copyright law to recognize style as a protectable element under certain conditions, particularly when a style is uniquely associated with a specific living artist \cite{jiang2023ai}.
(iv) \textit{Arms race.} However, LightShed \cite{foerster2025lightshed} demonstrates that perturbation-based protections like Glaze can be defeated with 99.98\% accuracy by a trained DNN, suggesting that the current generation of adversarial protection tools may not provide permanent solutions and that the protection--attack arms race will continue to escalate.
\end{addedblock}

\vspace{-3mm}
\subsection{Society}
From a societal perspective, as shown in Fig. \ref{Fig: influence} (c), the phenomenon of replication in visual diffusion models manifests in the duplication of human values, ideals, and even biases within generated images or videos. Much of the research in this area focuses on biases, as these can reinforce and amplify societal inequalities and discrimination. Different from the biases discussed in the Regulation subsection, we review papers from a societal perspective here.\par

\begin{addedblock}
\noindent\textbf{Unveiling societal bias replication.}
Several studies have systematically measured the extent to which diffusion models replicate societal biases.
\cite{luccioni2023stable} introduces a method for assessing social biases by analyzing how varying input prompts related to gender, ethnicity, and professions influence the diversity of generated images.
\cite{bianchi2023easily} demonstrates at large scale that ordinary prompts describing traits, occupations, and objects produce amplified demographic stereotypes in text-to-image models, with stereotypes persisting even after user-level counter-stereotype prompting.
\cite{cho2023dalleval} probes gender and skin-tone biases across professions, revealing that models learn specific demographic associations from web image-text training pairs.
\cite{zhang2023auditing,wu2023stable} specifically audit gender representation in text-to-image models, emphasizing how they reinforce gender stereotypes.
More recently, \cite{dinca2024openbias} proposes the first open-set bias detection pipeline using an LLM to hypothesize biases and a VQA model to verify them in generated images, discovering novel biases across Stable Diffusion 1.5, 2, and XL.
\cite{toxicitybias2025} further demonstrates that popular Stable Diffusion models respond to harmful prompts by generating content with troubling biases such as the disproportionate portrayal of certain racial groups in violent contexts.\par

\noindent\textbf{Understanding bias origins.}
The root causes of bias replication can be traced to multiple factors.
\cite{perera2023analyzing} highlights that diffusion models exacerbate biases present in their training data, with the effect depending on dataset size and composition.
\cite{dehdashtian2025oasis} introduces sociologically grounded stereotype scores and internal probing tools, showing that despite improved fidelity, newer models retain strong stereotypical predispositions; stereotypes are worse for nationalities with lower Internet footprints, tracing the root cause to training data distribution imbalances.
\cite{chen2024would} investigates feedback loops where generated images train future models; surprisingly, substituting real images with diffusion outputs does not uniformly amplify bias and can sometimes mitigate it, attributing this to generation artifacts that disrupt learned associations.
These findings collectively suggest that bias replication originates from the interaction between skewed training data distributions, CLIP-based text encoders that encode societal associations, and the high-fidelity generation capability that faithfully reproduces these patterns.\par

\noindent\textbf{Mitigation through debiasing.}
Several strategies have been proposed to reduce bias replication in diffusion models.
\cite{shen2024finetuning} proposes a distributional alignment loss with adjusted direct finetuning, markedly reducing gender, racial, and intersectional biases.
\cite{orgad2023editing} introduces TIME (Text-to-Image Model Editing), which corrects biased implicit assumptions (\eg, ``a CEO'' always generating a white male) by editing only $\sim$2\% of model parameters in under one second, without any retraining.
\cite{ddm2025debiasing} proposes learning latent representations that promote fairness without requiring predefined sensitive attributes.
Most recently, \cite{shi2025dissecting} identifies internal ``bias features'' within diffusion model architectures via mechanistic interpretability and directly manipulates them to achieve granular control over bias levels, bridging both understanding and mitigation within a unified framework.
\end{addedblock}

\subsection{Healthcare}
Visual diffusion models have significantly impacted the field of healthcare by enhancing the generation and analysis of medical images, which are critical tools in diagnosis, treatment planning, and research.

A primary way diffusion models assist in medical imaging is through the generation of synthetic images \cite{pan20232d,eschweiler2024denoising,peng2023generating}. These models can create realistic medical images, such as MRI scans or X-rays, from a dataset of existing images. This capability is particularly useful for training medical professionals, as it allows for the creation of diverse scenarios and conditions that might not be readily available in educational settings due to rarity or ethical concerns. Additionally, synthetic images can augment datasets used to train other machine learning models, improving their ability to recognize and diagnose conditions from real patient data. \par

Furthermore, diffusion models can enhance image quality and detail \cite{asgariandehkordi2023deep,xiang2022ddm,hu2022unsupervised}, which is vital in medical diagnostics where the clarity of an image can influence the accuracy of assessments made by radiologists. For example, diffusion models can refine images, improving resolution and contrast, or even reconstruct incomplete scans. This enhances the interpretability of medical images and assists in more accurate diagnosis and patient monitoring.\par

Moreover, visual diffusion models support the development of automated diagnostic tools \cite{kascenas2023role,wolleb2022diffusion,liang2023modality}. By generating high-quality, detailed images, these models aid in training algorithms that can detect anomalies such as tumors, fractures, or degenerative conditions. This speeds up the diagnostic process and helps in reducing human error by providing a consistent, objective analysis that can be used as a second opinion or to verify human-made diagnoses.\par

As shown in Fig. \ref{Fig: influence} (d), while visual diffusion models offer significant benefits in medical imaging, such as enhancing image quality and generating scarce datasets, these models also pose substantial risks due to their potential for replication. The replication phenomenon could lead to generated images being overly similar to real patient data, thus risking personal health information disclosure. In the following, we review papers that unveil, understand, and mitigate the replication phenomenon in the context of medical imaging.\par

\noindent\textbf{Unveiling replication in medical imaging.} Several studies have served as ``whistleblowers'' in highlighting these issues. For instance, the research \cite{dar2023investigating} reveals that 3D latent diffusion models are more prone to replicate original training images, affecting the model’s generalizability.
Another study \cite{akbar2023beware} compares diffusion models and GANs in synthesizing medical images, finding that diffusion models, compared to GANs, are more likely to replicate training images when generating 2D slices from 3D volumes, increasing the risk of patient re-identification.
Further research in \cite{dar2024unconditional} confirms the tendency of latent diffusion models to replicate data in an unconditional generation setting, suggesting that diffusion models might fail to prevent the disclosure of training data details even when not targeted for specific tasks. \par

\noindent\textbf{Understanding data and training factors.} The occurrence of replication in medical imaging, can be attributed mainly to two factors: the size of the original dataset and the number of training epochs.
Firstly, when diffusion models are trained on small datasets, there is a higher risk of replication, as the model has fewer examples from which to learn and generalize. Studies such as \cite{dar2023investigating} and \cite{usman2024brain} have highlighted that models trained on small datasets, such as those containing detailed scans for brain tumors, tend to produce synthetic images that too closely replicate the training images, reducing their utility and increasing privacy risks. Secondly, \cite{dar2024effect} reveals that over-training a model -- running too many epochs -- can lead to a situation where the diffusion models begin to precisely replicate the training patient data rather than generating diverse synthetic images. This occurs because excessive training on the same dataset reinforces the model’s exposure to and retention of specific data characteristics, thereby increasing the likelihood of producing identical or nearly identical images to those seen during training.\par

\noindent\textbf{Mitigation through privacy-preserving techniques.} There are effective solutions that can mitigate these risks and thus enhance the privacy and utility of synthetic data.
Firstly, the approach of \textit{privacy distillation}, as discussed in \cite{fernandez2023privacy} offers a robust method for safeguarding patient information. This technique involves training a diffusion model on real data to generate a synthetic dataset, which is then filtered to remove any potentially identifiable information. A second model is then trained exclusively on this sanitized dataset. Secondly, \textit{data augmentation} is another approach that can enhance the diversity of training datasets and reduce overfitting. By artificially expanding the dataset through transformations and variations of the original patient images, models are less likely to replicate \cite{dar2024effect,dar2023investigating}. \added{Thirdly, \textit{differentially private 3D synthesis} \cite{dp3dmedical2025} introduces controllable latent diffusion models under strict DP constraints for volumetric medical images, while comprehensive privacy assessments \cite{rethinkingprivacy2025} reveal that leakage risks extend throughout the entire deep learning pipeline, from data sharing to model deployment, necessitating holistic privacy-preserving strategies.}

\begin{addedblock}
\subsection{Emerging Applications}
\label{sec:emerging}
Beyond healthcare, the replication phenomenon poses unique risks in several rapidly growing application domains. We discuss three representative cases below.

\noindent\textbf{Metaverse and virtual environments.}
Diffusion models are increasingly used to generate 3D scenes, avatars, and virtual assets for metaverse platforms and game engines \cite{shen2024context}. This introduces novel replication risks:
(i) \textit{3D asset replication}\added{, \ie, the reproduction of copyrighted geometry or textures in generated 3D models}. As noted in Section~\ref{sec:video3d}, generated 3D assets may replicate the geometry or texture of copyrighted assets in training sets, enabling large-scale plagiarism of virtual goods whose commercial value can be substantial.
(ii) \textit{Real-space replication}\added{, \ie, the reproduction of identifiable physical locations in generated virtual environments}. Models trained on street-view or indoor imagery can replicate identifiable real-world locations (homes, offices, medical facilities) within virtual environments, raising both privacy (exposure of private spaces) and security concerns (reconnaissance).
(iii) \textit{Avatar identity replication}\added{, \ie, reproducing a real person's biometric identity in avatar form}. Diffusion models conditioned on facial images can replicate the biometric identity of real individuals in avatar form, enabling identity theft within virtual worlds. This is a concept-and-content dual replication: both the person's identity (concept) and their facial features (content) are replicated.

\noindent\textbf{Autonomous driving.}
Diffusion models are increasingly used to synthesize rare driving scenarios for training and testing autonomous vehicle (AV) systems \cite{zhao2024drivedreamer,yang2023unisim}. The replication phenomenon manifests here in safety-critical ways:
(i) \textit{Pedestrian identity replication}\added{, \ie, reproducing recognizable individuals from training data in synthetic traffic scenes}. Synthetic traffic datasets may replicate the appearance of real individuals photographed in training data, raising GDPR concerns and potentially causing discriminatory behavior if the model replicates demographic biases in pedestrian recognition.
(ii) \textit{Scenario memorization}\added{, \ie, reproducing the exact spatiotemporal dynamics of real driving incidents rather than generating novel scenarios}. When a model is trained on a limited set of accident recordings, it may replicate the exact spatiotemporal dynamics of real incidents rather than generalizing to novel scenarios \cite{gu2023memorization}. This makes the synthetic dataset less diverse than intended and may give a false sense of safety coverage.
(iii) \textit{Geographic location replication}\added{, \ie, reproducing identifiable roads, intersections, or landmarks that reveal proprietary training routes}. Models trained on geolocation-tagged datasets may replicate identifiable road segments, intersections, or landmarks, enabling inference of the geographic distribution of the AV company's proprietary training routes.

\noindent\textbf{Education and scientific research.}
Diffusion models are increasingly used to generate illustrative figures, data visualizations, and synthetic experimental images in educational materials and scientific publications. The replication phenomenon raises concerns in this domain:
(i) \textit{Scientific image fabrication.} Generated images intended as novel illustrations may inadvertently replicate figures from published papers in the training data, leading to unintentional plagiarism or fabrication allegations.
(ii) \textit{Training data leakage in synthetic datasets.} When diffusion models generate synthetic data for educational or research purposes (\eg, augmenting small datasets for student projects), replicated training samples may introduce bias or violate the privacy of original data subjects.
(iii) \textit{Erosion of academic integrity.} The ease of generating realistic images undermines the evidentiary value of visual data in scientific publications, as reviewers and readers cannot distinguish genuine experimental results from AI-generated replications of prior work.

Mitigating replication in these emerging domains requires domain-specific adaptations of the general strategies, particularly data minimization, differential privacy, and machine unlearning, as well as the development of domain-specific benchmarks. The VideoShield framework \cite{hu2025videoshield} provides a first step toward regulation in the video domain through watermark-based provenance tracking, and \cite{t2vunlearning2025} extends concept erasure to text-to-video models.
\end{addedblock}

\section{Challenges and Future Directions}
\label{Challenges}
After reviewing the replication issues in visual diffusion models, including unveiling, understanding, mitigation, and its influence in the real world, this section will discuss the current challenges and future directions in this field.
\subsection{Specialized Visual Copy Detection}
\added{Unlike generic image retrieval, specialized visual copy detection refers to detection models specifically designed to identify diffusion-generated replications across content, style, and concept levels.}
Currently, many research efforts \cite{somepalli2023diffusion,bralios2024generation,rahman2024frame,zhou2023copyscope,aboutalebi2023deepfakeart,wu2024cgi} focus on the analysis of \textit{replicated content} because this level of replication aligns well with human perceptions of similarity. However, these methods predominantly rely on existing feature extraction models, such as SSCD \cite{pizzi2022self} and CLIP \cite{radford2021learning}, which are not specifically designed for diffusion-based replication. SSCD \cite{pizzi2022self}, for instance, only learns invariance against image transformations, while CLIP \cite{radford2021learning} is developed primarily for natural images. Consequently, these models often fail to detect some forms of replicated content generated by diffusion models. As a result, the analysis of replicated content by visual diffusion models tends to be both inaccurate and biased.\par

Future efforts could focus on creating a new dataset that includes images and videos featuring various types of replicated content generated by visual diffusion models. This dataset would then be used to train specialized visual copy detection models. By employing these models, subsequent analysis is expected to become both more accurate and fairer.

\subsection{In-context Similarity Retrieval}
\added{In-context similarity retrieval refers to adapting a single foundational retrieval model to detect different types of replication (content, style, concept) by providing contextual examples at inference time, without retraining.}
Replication in visual diffusion models manifest across various dimensions, including gender \cite{anand2023identifying}, culture \cite{struppek22homoglyphs}, racial aspects \cite{luccioni2023stable}, NSFW content \cite{schramowski2023safe}, copyrighted images \cite{somepalli2023understanding}, patient information \cite{kazerouni2023diffusion}, photos of politicians or celebrities \cite{chen2023text}, and artistic styles \cite{wang2024AnyPattern}. Current approaches to unveiling this phenomenon typically utilize a spectrum of feature extraction models or purpose-specific models. Although these methods can be effective, they come with significant disadvantages: (1) Selecting suitable models for practical deployment is time-consuming; (2) labeling new datasets and training specialized models are costly; and (3) the models, once trained, often lack generalizability to other contexts, thereby increasing the overall costs of practical applications.

In light of these challenges, introducing in-context learning to this area presents a promising direction for future development. In-context learning, a paradigm where a single foundational model adapts to a variety of tasks based on the context provided during inference, eliminates the need for multiple specialized models. Specifically, for in-context similarity retrieval, this approach could enable the foundational model to dynamically adjust the feature extraction process based on specific concerns, such as gender biases, racial characteristics, or copyrighted content. This is achieved by simply presenting relevant contextual examples, which allows adjusting without the need for retraining.
This methodology eliminates the need for selecting/training specialized models and significantly increases generalizability.

\subsection{Robust Mitigation}
The current mitigation methods often fail to successfully solve the replication problems. For instance,
\begin{itemize}[leftmargin=*]
\item Even after deduplicating images and captions in the dataset, visual diffusion models can still generate samples similar to data points from the training set;
\item Malicious researchers can easily develop AI methods to bypass training data protection strategies\added{, as demonstrated by LightShed~\cite{foerster2025lightshed} defeating Glaze with 99.98\% accuracy};
\item Visual diffusion models equipped with machine unlearning methods can still generate concepts that have already been erased\added{, with recent work~\cite{george2025illusion} showing that erased concepts re-emerge under slight perturbations};
\item Prompt perturbation methods cannot always generate images that align well with the prompt while simultaneously avoiding the creation of copyrighted content.
\end{itemize}
Therefore, in the future, researchers can (1) focus on enabling visual diffusion models to learn only the semantic content from any training sample, rather than its specific details; (2) develop irremovable protection mechanisms for images and videos; \added{(3) explore strategies that combine deduplication, differential privacy, and unlearning at different stages of the pipeline;} and (4) enhance prompt engineering techniques to ensure that the generated content not only avoids legal pitfalls but also more accurately reflects users' intent.

\subsection{Unified Benchmarks}

In the context of computer vision, benchmarks serve as crucial tools for measuring and comparing the performance of various algorithms across standardized tasks and datasets. Although research on replication is flourishing, researchers often work in isolation, leading to inconsistencies in evaluation. Future research may focus on building unified benchmarks for comparing algorithms in unveiling, understanding, and mitigating replication.\par

\noindent\textbf{Unveiling.} The benchmarks for unveiling may evaluate the accuracy of current methods. For instance, many membership inference methods claim that they can nearly judge with 100\% accuracy whether a visual diffusion model was trained on one specific image or video. However, \added{as shown in Table~\ref{tab:mia},} this may not be the case in the real world setting\added{, where black-box AUC drops to near-random levels}. Therefore, building a benchmark to compare which membership inference attack is most effective is essential.

\noindent\textbf{Understanding.} Currently, all the understanding of replication phenomenon seems to be reasonable and correct. However, each understanding on replication may only captures a part of the entire complexity. Building a unified benchmark for understanding is challenging, yet it provides a valuable measure of the applicability of different interpretations.

\noindent\textbf{Mitigation.} The benchmark for mitigation may include assessing how successful the data optimization strategies and unlearning methods are. Specifically, researchers may evaluate what proportion of the training data is replicated using different protection or unlearning strategies.

\subsection{New Regulation}
As AI's capability to mimic human characteristics in creative outputs grows, there is an increasing need for transparency in AI's role in content creation. This transparency is crucial for addressing copyright claims and enhancing public understanding. Current initiatives focus primarily on the necessity of disclosing AI involvement in registered works. Furthermore, the development and training of AI systems often involve using large volumes of data, some of which includes copyrighted material. This practice has sparked concerns over potential copyright infringement, an issue that remains legally ambiguous. \added{As discussed in Section~\ref{Medical} and Table~\ref{tab:regulation}, the regulatory landscape varies significantly across jurisdictions, with the EU mandating opt-out mechanisms while the US relies on case-by-case fair use analysis.} Consequently, there is a pressing need for updated regulations or clarifications on existing copyright exceptions to accommodate the complexities introduced by AI. Jurisdictions worldwide are beginning to consider such amendments, but the global landscape is still uneven and undergoing transition. \added{Emerging domains such as metaverse, autonomous driving, and education (Section~\ref{sec:emerging}) further complicate the regulatory picture, as domain-specific replication risks require tailored legal frameworks.}

\end{document}